\documentclass{article}

\usepackage[main,final,nonatbib]{neurips_2026}

\usepackage[utf8]{inputenc} % allow utf-8 input
\usepackage[T1]{fontenc}    % use 8-bit T1 fonts
\usepackage{hyperref}       % hyperlinks
\usepackage{url}            % simple URL typesetting
\usepackage{booktabs}       % professional-quality tables
\usepackage{amsfonts}       % blackboard math symbols
\usepackage{nicefrac}       % compact symbols for 1/2, etc.
\usepackage{microtype}      % microtypography
\usepackage[table]{xcolor}  % colors

\hypersetup{hidelinks}

\usepackage[style=authoryear, natbib, giveninits=true, uniquename=init, maxbibnames=100, dashed=false, date=year]{biblatex}
\let\cite\citep

\AtBeginBibliography{
  \DeclareNameAlias{author}{family-given}
  \DeclareDelimFormat{multinamedelim}{\addsemicolon\space}
  \DeclareDelimFormat{finalnamedelim}{\addsemicolon\space\bibstring{and}\space}
}
\renewbibmacro*{finentry}{%
  \iffieldundef{annotation}
    {}
    {\setunit{\addperiod\space}%
     \printfield{annotation}}%
  \finentry}

\usepackage[shortcuts=ac]{glossaries-extra}
\setabbreviationstyle[acronym]{long-short}
\newacronym{bce}{BCE}{binary cross-entropy}
\newacronym{gnn}{GNN}{graph neural network}
\newacronym{mlp}{MLP}{multi-layer perceptron}
\newacronym[longplural={Markov decision processes}]{mdp}{MDP}{Markov decision process}
\newacronym{rgcn}{R-GCN}{relational graph convolutional network}
\newacronym{rim}{RIM}{recurrent independent mechanism}
\newacronym{dias}{DIAS}{Differentiable Induction of Action Schemas}
\glsunset{dias}
\newacronym{mcts}{MCTS}{Monte-Carlo tree search}

\glsdisablehyper

\usepackage{mleftright}
\mleftright

\usepackage{siunitx}
\usepackage{listings}
\usepackage[capitalise, noabbrev]{cleveref}
\AddToHook{cmd/appendix/before}{%
  \crefalias{section}{appendix}%
  \crefalias{subsection}{appendix}
  \crefalias{subsubsection}{appendix}
}

\usepackage{caption}
\makeatletter
\renewenvironment{table}
  {\setlength{\abovecaptionskip}{\@neuripsabovecaptionskip}%
   \setlength{\belowcaptionskip}{\@neuripsbelowcaptionskip}%
   \@float{table}}
  {\end@float}
\makeatother

\usepackage{svg}
\usepackage{tabularx}
\usepackage{mathtools}
\usepackage[inline]{enumitem}
\usepackage{xspace}

\newcommand*{\tuple}[2][]{#1\langle #2 #1\rangle}
\newcommand*{\set}[2][]{#1\{ #2 #1\}}
\newcommand*{\abs}[2][]{#1\lvert #2 #1\rvert}
\newcommand*{\norm}[2][]{#1\lVert #2 #1\rVert}
\newcommand*{\bmat}[1]{\begin{bmatrix} #1 \end{bmatrix}}
\newcommand*{\given}[1][]{\;#1|\;}

\newcommand{\sift}{\textnormal{SIFT}\xspace}
\newcommand{\synth}{\textnormal{SYNTH}\xspace}
\newcommand{\strips}{\textnormal{STRIPS}\xspace}
\newcommand{\stripsp}{\textnormal{STRIPS+}\xspace}

\title{Differentiable Learning of Lifted Action Schemas \\ for Classical Planning}

\author{%
  Jonas Reiter\\
  RWTH Aachen University\\
  \texttt{jonas.reiter@ml.rwth-aachen.de} \\
  \And
  Jakob Elias Gebler\\
  RWTH Aachen University\\
  \texttt{jakob.gebler@ml.rwth-aachen.de}\\
  \And
  Hector Geffner\\
  RWTH Aachen University\\
  \texttt{hector.geffner@ml.rwth-aachen.de}\\
}

\begin{document}

\maketitle

\begin{abstract}
  Classical planners can effectively solve very large deterministic MDPs represented in STRIPS or PDDL where states are sets of atoms over objects and relations, and lifted action schemas add or delete these atoms. This compact representation yields strong search heuristics and provides an ideal setting for structural generalization, since lifted relations and action schemas give rise to infinitely many domain instances. A central challenge is to learn these relations and action schemas from data, and recent approaches have addressed this problem using different types of observations. In this work, we develop a novel neural network architecture for learning action schemas from traces where states are fully observed but action arguments are unobserved. The problem is a simplification but an important step towards learning planning domains from sequences of images and action labels, and we aim to solve this simplification in a nearly perfect manner. The challenge lies in learning the action schemas while simultaneously identifying the action arguments from observed state changes. Our approach yields a robust differentiable component that can then be integrated into larger neuro-symbolic models. We evaluate the architecture on various planning domains, where the learned lifted action schemas must recover the ground-truth structure. Additionally, we report experiments on robustness to observation noise, on component ablations, and on a variation related to slot-based dynamics models.
\end{abstract}

\section{Introduction}

Classical planners can find goal-directed plans in very large deterministic \acp{mdp}. They solve these problems from scratch, without training, by automatically deriving heuristic information to guide search from STRIPS-like representations in which states are sets of atoms over a fixed set of relations and actions add and delete atoms provided that certain preconditions hold \cite{geffner_concise_2013,ghallab2025acting}. These reachability problems involve huge state spaces that are beyond the scope of exploration methods in reinforcement learning and \ac{mcts}.

A key challenge in classical planning is to learn state relations and action schemas from observations rather than hand-crafting them. Recent approaches address this problem in different forms. Traditional symbolic methods learn from traces consisting of full states and full actions, sometimes under partial observability or noise \cite{zhuo2013action,aineto2019learning,le2024learning,bachor2024learning,paolo:2025}. More recently, it has been shown that state relations and action schemas can be learned effectively and correctly from pure action traces alone \cite{sift}. A limitation of these settings is that actions are assumed to reveal all of their arguments, even though some arguments are not truly observable and others are merely artifacts of the \strips language. For example, the action of picking up a block $x$ in \strips is written as $\mathtt{unstack}(x,y)$, where $y$ is the block on which $x$ is currently placed. The only reason $y$ appears explicitly is that the action adds the atom $\mathtt{clear}(y)$. In other planning languages, such as \stripsp, the action can instead be written as $\mathtt{unstack}(x)$, with the value of $y$ determined uniquely by the precondition $\mathtt{on}(x,y)$. The \synth algorithm learns action schemas from traces containing full states and actions with the minimal number of arguments needed to determine all remaining ones \cite{jansen_learning_2025}. Likewise, a recent SAT-based approach learns action schemas from full states and action names alone, without assuming knowledge of the action arguments or their arity \cite{balyo:2024}. For this setting, the authors show that the problem is computationally as hard as establishing an isomorphism between two graphs.

In this work, we propose \acs{dias} (\acl{dias}),\footnote{Code:
\url{https://github.com/jonaslreiter/DIAS}} a novel neural architecture for learning action schemas from transitions in which states are fully observed, possibly with noise, but action arguments are not. We address the same isomorphism-hard problem as above, but through gradient descent rather than repeated SAT calls. The potential advantages include robustness to noise and the possibility of replacing full symbolic states with other types of observations, such as images. The learned schemas also scale to larger problem instances unchanged. More generally, the problem we study is a simplification of the harder and more realistic task of learning a planning domain from sequences of images and action labels \cite{xi_neurosymbolic_2024,xi_learning_2026}. Rather than optimizing relative performance on that broader task, however, we aim for near-perfect performance on this simplified setting, which remains a necessary and challenging component in its own right. Data without action arguments arise when a system logs operations from a fixed set of actions with no notion of objects. Moreover, some implicit arguments would only be determined once the schema is fixed. We also consider two variants of the learning problem in which some or all action arguments are observed.

\begin{figure}[t]
  \centering
  \includesvg[width=\textwidth]{figures/approach-figure.svg}
  \caption{Overview of the \acs{dias} architecture. A transition $\tuple{s, s'}$ is encoded as a graph $G$. A \acs{gnn} computes object embeddings $K$, which are matched with object queries $Q^a$ for action $a$. Together with slot weights $w^a$, this yields an object selection $\tilde{S}^a$. Lifted preconditions and effects $P^a_{\mathrm{pre}^+}$, $P^a_{\mathrm{pre}^-}$, $P^a_\mathrm{add}$, and $P^a_\mathrm{del}$ are then grounded with $\tilde{S}^a$ into adjacency-matrix space as $\tilde{P}^a_{\dots}$. Given the adjacency matrix $A$ of state $s$, grounded preconditions are checked and effects are applied to predict the successor-state adjacency matrix $\hat{A}'$. Training minimizes a prediction loss between $\hat{A}'$ and $A'$ while additionally maximizing and minimizing lifted preconditions and effects, respectively. After training, the action schema for each action $a$ can be derived from its lifted parameters.}
  \label{fig:approach-overview}
\end{figure}

The proposed \ac{dias} architecture is shown in \cref{fig:approach-overview} and operates in two stages. First, a \ac{gnn} takes as input a graph defined by the atoms in state $s_t$ at time $t$ and their changes in the next state $s_{t+1}$ after action $a_t$. It computes embeddings for each object in the state that act as keys $K$. For each action name $a$, we maintain a set of learnable query embeddings $Q^a$ and select a list $o_1, \ldots, o_k$ of $k$ objects into slots in $\tilde{S}^a$ via attention between $K$ and $Q^a$. In addition, learnable slot probabilities $w^a$ determine whether a slot is active. Second, for each action name $a$, we maintain four learnable probabilities for every lifted atom $p(x_i,x_j)$, with $i,j \in [1,k]$, in $P^a_{\mathrm{pre}^+}$, $P^a_{\mathrm{pre}^-}$, $P^a_\mathrm{add}$, and $P^a_\mathrm{del}$. These encode whether the atom is a lifted positive or negative precondition, an add effect, or a delete effect of the action. The object selection $\tilde{S}^a$ grounds the lifted action in the state, where preconditions must hold in $s_t$ and effects must predict the next state $s_{t+1}$. Prediction errors are minimized with a suitable loss function.

The selection of the objects that bind action arguments, together with the localized effects of actions on those objects, connects \ac{dias} to a number of slot-based dynamic models in deep learning, such as \acp{rim} and related approaches \cite{rim,rim1,rim2,rim3}. In these models, slots play the role of state variables that interact sparsely. Two important differences from \strips dynamics are that \strips effects are state independent and therefore less expressive, yet the learned \strips action schemas generalize in ways that slot-based or variable-based dynamics cannot. The reason is that in domains such as \emph{Blocksworld}, changing the number of objects changes not only the number of state variables, but also their set of possible values. This does not happen in \strips, where the state variables, namely atoms, are Boolean. We also consider a \ac{rim}-like architecture in our setting in which the \strips effects are replaced by a \ac{mlp} that takes the values of atoms over the selected objects at time $t$ and predicts their effects. The resulting dynamic model is more expressive than \strips because it handles state-dependent effects, and it generalizes to larger numbers of objects in the same way, but learning is less effective and does not yield symbolic action schemas that support effective planning.

The remainder of the paper is organized as follows. We cover the necessary background and notation in \cref{sec:preliminaries}, formulate the problem setting in \cref{sec:problem-formulation}, introduce \ac{dias} in detail in \cref{sec:approach}, report experiments in \cref{sec:experiments}, and conclude with limitations in \cref{sec:conclusions}. A more detailed discussion of related work is provided in \cref{app:related-work}, and ablations and sensitivity analyses in \cref{app:ablation-studies}.

\section{Preliminaries}
\label{sec:preliminaries}

We review STRIPS and the STRIPS+ variant for modeling classical planning domains.

\subsection{STRIPS}

In classical planning, a STRIPS problem~$P$ is defined as a pair $P = \tuple{D, I}$ consisting of a domain~$D$ and instance information~$I$~\citep{geffner_concise_2013}. Here, $D = \tuple{\mathcal{P}, \mathcal{A}}$ is a lifted domain description containing a set of predicates $\mathcal{P} = \set{p_i}_{i=1}^R$ and a set of action schemas $\mathcal{A} = \set{a_i}_{i=1}^A$. Each predicate $p \in \mathcal{P}$ takes a set of arguments $p(y_1, \dots, y_{\abs{p}})$, or, more briefly, $p(\vec{y})$, where $\abs{p}$ denotes the arity of predicate~$p$. We denote by $\mathcal{P}_k = \set{p \in \mathcal{P} \given \abs{p}=k}$ the subset of predicates with arity~$k$. Each action schema $a \in \mathcal{A}$ similarly takes a set of arguments $a(x_1, \dots, x_{\abs{a}})$, or, more briefly, $a(\vec{x})$, where $\abs{a}$ denotes the arity of action schema~$a$. An action schema is defined by its set of preconditions, add effects, and delete effects as $a(\vec{x}) = \tuple{\mathrm{Pre}(a), \mathrm{Add}(a), \mathrm{Del}(a)}$. These sets consist of lifted atoms $p(x_1, \dots, x_{\abs{p}})$ over the action schema arguments.

An instance $I = \tuple{\mathcal{O}, s_0, g}$ consists of a set of constants or objects $\mathcal{O} = \set{o_i}_{i=1}^O$, an initial state~$s_0$, and a goal description~$g$. A predicate $p(o_1, \dots, o_{\abs{p}})$, with objects~$\vec{o}$ assigned to its arguments~$\vec{y}$, is called a ground atom. Then $s_0$ is the set of all ground atoms true in the initial state, and $g$ is a set of ground atoms that have to be true in every goal state.

Applying a ground action $a(o_1, \dots, o_{\abs{a}})$, with objects~$\vec{o}$ assigned to action arguments~$\vec{x}$, induces a transition from a current state~$s$ to a next state~$s'$ in the state space~$\mathcal{S}$. The grounded preconditions $\mathrm{Pre}(a(\vec{o}))$ must hold in $s$ for $a(\vec{o})$ to be applicable. Action execution then yields $s' = ((s \setminus \mathrm{Del}(a(\vec{o}))) \cup \mathrm{Add}(a(\vec{o})))$.

\subsection{STRIPS+}

In STRIPS, the action arguments $\tuple{x_1, \dots, x_{\abs{a}}}$ must contain every variable that appears in a lifted atom $p(x_1, \dots, x_{\abs{p}})$ in the preconditions $\mathrm{Pre}(a)$ or effects $\mathrm{Add}(a)$ and $\mathrm{Del}(a)$. \Citet{jansen_learning_2025} introduce STRIPS+ to relax this requirement. In STRIPS+, lifted preconditions and effects may use an additional set of variables $\vec{z}$ that is disjoint from the explicit variables $\vec{x}$ and does not appear in the action arguments. The only requirement is that the value of every variable in $\vec{z}$ is uniquely determined by the explicit action arguments $\vec{x}$ together with the action preconditions. STRIPS+ can therefore express STRIPS domains while using fewer action arguments.

Intuitively, this allows for more natural action descriptions. For example, when moving a block in \emph{Blocksworld}, the STRIPS action \texttt{move(?block ?from ?to)} can be simplified to \texttt{move(?block ?to)} in STRIPS+, since the source location is uniquely determined by the precondition \texttt{on(?block ?from)}. In our approach, we consider input traces consisting of full states together with STRIPS+ actions, full STRIPS actions, or action names only.

\section{Problem formulation}
\label{sec:problem-formulation}

We are given a set of $N$ state-transition triplets $\set{\tuple{s, a, s'}_i}_{i=1}^N$. All transitions are applicable and sampled under an arbitrary stochastic policy from one or more instances~$I$ of a planning domain~$D$. They do not need to form sequences and may be fragmentary or unordered. The states $s$ and $s' \in \mathcal{S}$ are grounded STRIPS state descriptions. We consider three cases for the action label~$a$:
\begin{enumerate*}[label=(\roman*)]
  \item full STRIPS actions, e.g., \texttt{move(o\textsubscript{3},o\textsubscript{7},o\textsubscript{4})},
  \item STRIPS+ actions, e.g., \texttt{move(o\textsubscript{3},o\textsubscript{4})}, or
  \item action names only, e.g., \texttt{move}.
\end{enumerate*}

Our goal is to learn the lifted preconditions and effects $\tuple{\mathrm{Pre}(a), \mathrm{Add}(a), \mathrm{Del}(a)}$ of each action schema $a \in \mathcal{A}$ in the hidden lifted domain. A learned domain is considered equivalent to the original domain if, for any state~$s$ from any instance~$I$, it generates the same set of successor states~$s'$. This entails soundness, i.e., all generated successors are true successors, and completeness, i.e., all true successors are generated. We do not aim to recover the original schemas syntactically. A learned schema may order and name its arguments differently or contain redundant arguments and preconditions.

\paragraph{Assumptions.}
We do not consider conditional effects, numerical features, or other advanced PDDL features, only pure STRIPS domains with typing and negative preconditions. We also adopt the injective action binding assumption~\citep{juba_safe_2021}, ruling out domains in which a single object~$o$ can be bound to multiple action arguments. Finally, we do not consider domains with predicates of arity other than $1$ or $2$, and we assume that all domain predicates are known. On the other hand, we do not assume that the action arities are known.

\section{Approach}
\label{sec:approach}

Our goal is to learn lifted preconditions and effects for each action $a(\vec{x})$. We do so by considering individual transitions $\tuple{s, a, s'}$. \Ac{dias} selects the arguments $\vec{x}$ of $a$, ensures that learned preconditions are met, and applies learned effects to compute a successor-state prediction $\hat{s}'$. All stages use soft and differentiable operations to allow for end-to-end gradient-based optimization. The complete approach is depicted in \cref{fig:approach-overview}, and the individual stages are introduced below. \Cref{app:walkthrough} follows one \emph{Blocks-3} transition through all stages.

\subsection{Selection of action arguments}
\label{sec:argument-selection}

In general, we do not assume knowledge of the objects $\vec{o}$ to assign to the action arguments $\vec{x}$. Instead, we infer them from states $s$ and $s'$ using embeddings computed by a \ac{gnn}.

The state $s$ is a set of ground atoms $p(\vec{o})$ from which we construct a directed graph $G = \tuple{V, E}$. The nodes $V$ are given by the set of objects $\mathcal{O}$, and each ground atom $p(o_i, o_j)$ true in $s$ represents a typed edge $\tuple{p, \tuple{o_i, o_j}}$ of type $p$ between objects $o_i$ and $o_j$. We additionally determine $p'_\mathrm{add}(\vec{o})$ and $p'_\mathrm{del}(\vec{o})$ as the sets of ground atoms added and deleted when transitioning from state $s$ to $s'$. The respective typed edges $\tuple{p'_\mathrm{add}, \tuple{o_i, o_j}}$ and $\tuple{p'_\mathrm{del}, \tuple{o_i, o_j}}$ are added to the set of edges $E$.

Note that we limit ourselves to unary and binary predicates. Unary atoms $p(o_i)$ are represented by self-loops $\tuple{p, \tuple{o_i, o_i}}$. Object types are included analogously as unary atoms $t(o_i)$ and edges $\tuple{t, \tuple{o_i, o_i}}$. If any action arguments $x_k = o_i$ are given, we mark them in the graph via self-loops $\tuple{a^k, \tuple{o_i, o_i}}$, where $a^k$ denotes an edge type for the $k$-th argument of action $a$. Finally, we add the reverse edge $\bar{e}$ for every edge $e \in E$ to facilitate message passing.

The resulting graph $G$, with random initial embeddings redrawn in every forward pass, is passed through a \ac{rgcn}~\citep{schlichtkrull_modeling_2018}. We denote the final $d$-dimensional node feature vectors for the objects as keys $K \in \mathbb{R}^{O \times d}$. To assign objects $o_i$ to action arguments $x_k$, each lifted action $a$ has a set of $d$-dimensional learnable query vectors $Q^a \in \mathbb{R}^{M \times d}$. Here, $M$ is the number of slots, a hyperparameter and upper bound on the action arity that can be represented. We compute how well each object's key $K_i$ matches each slot's query $Q^a_k$ through a correspondence matrix $C^a \in \mathbb{R}^{M \times O}$ akin to scaled dot-product attention~\citep{vaswani_attention_2017} as
\begin{equation}
  C^a = \frac{Q^a K^\top}{\sqrt{d}}
  \text{.}
\end{equation}

We then utilize a rectangular version of the Sinkhorn algorithm~\citep{brun_differentiable_2022} to compute a soft assignment from objects to mutually exclusive slots. The result $S^a = \operatorname{Sinkhorn\_D1D2}(C^a) \in (0, 1)^{M \times O}$ is a row-stochastic matrix with row sums $=1$ and column sums $\leq 1$. Each entry $S^a_{k,i}$ represents the probability that object $o_i$ is assigned to slot $k$.

Additionally, we introduce learnable parameters $w^a \in \mathbb{R}^M$. With a sigmoid activation, they represent the activation probability of each slot. We scale each slot's assignment distribution with the respective probability to obtain the selection matrix
\begin{equation}
  \tilde{S}^a = \operatorname{diag}(\sigma(w^a)) S^a
  \text{.}
\end{equation}

\subsection{Application of effects}

Each atom $p(x_1, \dots, x_k)$ of the lifted action schema $a(\vec{x})$ is either
\begin{enumerate*}[label=(\roman*)]
  \item not included in effects,
  \item an add effect, or
  \item a delete effect.
\end{enumerate*}
We therefore model the lifted effects as a learnable parameter matrix $P^a_\mathrm{eff} \in \mathbb{R}^{R \times M \times M \times 3}$ for each action name $a$, where $R$ is the number of predicates and $M$ the number of slots. For unary predicates, only the diagonal entries are learnable and off-diagonal entries are fixed to $\bmat{0 & -\infty & -\infty}$. Applying the softmax function over the last dimension yields the probabilities for each atom being an add or delete effect as
\begin{equation}
  \label{eq:effect-probabilities}
  P^a_{\mathrm{add},r,i,j} = \operatorname{softmax} \bigl( P^a_\mathrm{eff} \bigr)_{r,i,j,2} = \frac{\exp(P^a_{\mathrm{eff},r,i,j,2})}{\sum_{l=1}^3 \exp(P^a_{\mathrm{eff},r,i,j,l})}
  \quad \text{and} \quad
  P^a_{\mathrm{del}} = \operatorname{softmax} \bigl( P^a_\mathrm{eff} \bigr)_{\dots,3}
  \text{.}
\end{equation}

Given a selection matrix $\tilde{S}^a$, we then project the lifted effect probabilities $P^a_\mathrm{add}$ and $P^a_\mathrm{del} \in (0, 1)^{R \times M \times M}$ onto grounded atoms in adjacency matrix space, yielding $\tilde{P}^a_\mathrm{add}$ and $\tilde{P}^a_\mathrm{del} \in (0, 1)^{R \times O \times O}$ via
\begin{equation}
  \tilde{P}^a_\mathrm{add} = (\tilde{S}^a)^\top P^a_\mathrm{add} \tilde{S}^a
  \qquad \text{and} \qquad
  \tilde{P}^a_\mathrm{del} = (\tilde{S}^a)^\top P^a_\mathrm{del} \tilde{S}^a
  \text{.}
\end{equation}

For each grounded atom, these describe the probabilities of it being an add or delete effect. Given the adjacency matrix $A \in \set{0, 1}^{R \times O \times O}$ of state $s$ and the projected effect probabilities, we compute a prediction of the successor-state adjacency matrix $\hat{A}' \in (0, 1)^{R \times O \times O}$ via
\begin{equation}
  \label{eq:effect-application}
  \hat{A}' = A + p_\mathrm{pre} \cdot \bigl( (1 - A) \odot \tilde{P}^a_\mathrm{add} - A \odot \tilde{P}^a_\mathrm{del} \bigr)
  \text{.}
\end{equation}

Here, the scalar $p_\mathrm{pre} \in (0, 1)$ expresses the probability of all preconditions being fulfilled. If they are not fulfilled, no changes will be applied. The computation of $p_\mathrm{pre}$ is detailed in the next section. Gating the effects in this way means that a transition without changes is explained as well by unfulfilled preconditions as by effects that already hold. This would allow learning from data including unlabeled inapplicable actions, but a learned precondition is then only refuted by transitions that change the state (see \cref{sec:main-results}).

\subsection{Evaluation of preconditions}

Analogously to effects, each atom $p(x_1, \dots, x_k)$ of the lifted action schema $a(\vec{x})$ is either
\begin{enumerate*}[label=(\roman*)]
  \item not included in preconditions,
  \item a positive precondition, or
  \item a negative precondition.
\end{enumerate*}
We therefore introduce learnable parameters $P^a_\mathrm{pre} \in \mathbb{R}^{R \times M \times M \times 3}$ for each action name $a$ and apply a softmax along the last dimension to obtain $P^a_{\mathrm{pre}^+}$ and $P^a_{\mathrm{pre}^-} \in (0, 1)^{R \times M \times M}$. We also project them into adjacency matrix space for a given selection $\tilde{S}^a$ with
\begin{equation}
  \tilde{P}^a_{\mathrm{pre}^+} = (\tilde{S}^a)^\top P^a_{\mathrm{pre}^+} \tilde{S}^a
  \qquad \text{and} \qquad
  \tilde{P}^a_{\mathrm{pre}^-} = (\tilde{S}^a)^\top P^a_{\mathrm{pre}^-} \tilde{S}^a
  \text{.}
\end{equation}

For each grounded atom, $\tilde{P}^a_{\mathrm{pre}^+}$ and $\tilde{P}^a_{\mathrm{pre}^-}$ describe the probabilities of it being a positive or negative precondition. The failure cases are false atoms for positive preconditions and true atoms for negative preconditions. We multiply their contrary probabilities for each atom and evaluate their conjunction for the complete state's precondition fulfillment probability as
\begin{equation}
  p_\mathrm{pre} = \smashoperator{\prod_{r,i,j=1}^{R, O, O}} {\bigl( \bigl( 1-\tilde{P}^a_{\mathrm{pre}^+} \odot (1 - A) \bigr) \odot \bigl( 1 - \tilde{P}^a_{\mathrm{pre}^-} \odot A \bigr) \bigr)_{r,i,j}}
  \text{.}
\end{equation}

In practice, this product is very close to zero for a randomly initialized $P^a_\mathrm{pre}$, which prevents any application of learned effects in \cref{eq:effect-application}. We therefore begin with the geometric mean and schedule the root degree, with $\tau$ exponentially decaying from $1$ to $0$, so as to approach the product, computing
\begin{equation}
  \label{eq:pre-factor-schedule}
  p_\mathrm{pre} = \Biggl( \smashoperator[r]{\prod_{r,i,j=1}^{R, O, O}} {\bigl( \bigl( 1-\tilde{P}^a_{\mathrm{pre}^+} \odot (1 - A) \bigr) \odot \bigl( 1 - \tilde{P}^a_{\mathrm{pre}^-} \odot A \bigr) \bigr)_{r,i,j}} \Biggr)^{\tfrac{1}{\tau \cdot R \cdot O^2 + (1-\tau)}}
  \text{.}
\end{equation}

\subsection{Loss formulation}
\label{sec:loss-formulation}

Our primary objective and learning signal is the correct prediction of the successor state $s'$ from state $s$ and action name $a$. This is done via \cref{eq:effect-application} and therefore includes the selection of action arguments and the evaluation of preconditions through $p_\mathrm{pre}$. We compute a \ac{bce} loss with sum aggregation between the predicted and true successor adjacency matrices $\hat{A}'$ and $A'$ as
\begin{equation}
  L_\mathrm{adj} = \operatorname{BCE} \bigl( \hat{A}', A' \bigr) = -\smashoperator{\sum_{r,i,j=1}^{R, O, O}} \bigl( A'_{r,i,j} \cdot \log \bigl( \hat{A}'_{r,i,j} \bigr) + \bigl( 1 - A'_{r,i,j} \bigr) \cdot \log \bigl( 1 - \hat{A}'_{r,i,j} \bigr) \bigr)
  \text{.}
\end{equation}

Additionally, we seek the minimal effects that still lead to correct predictions and the most restrictive preconditions that still keep valid state transitions applicable. To achieve this, we consider the first cases corresponding to inactive effects and preconditions, given by
\begin{equation}
  P^a_{\mathrm{no\_eff}} = \operatorname{softmax} \bigl( P^a_\mathrm{eff} \bigr)_{\dots,1}
  \qquad \text{and} \qquad
  P^a_{\mathrm{no\_pre}} = \operatorname{softmax} \bigl( P^a_\mathrm{pre} \bigr)_{\dots,1}
  \text{.}
\end{equation}

For minimal effects, we maximize the lifted inactive case $P^a_{\mathrm{no\_eff}}$ through a \ac{bce} loss with target $1$. For maximal preconditions, we do the same to maximize $(1 - P^a_{\mathrm{no\_pre}})$, but additionally weight each entry by the activation probabilities of its slots. Thus we compute
\begin{equation}
  L_\mathrm{eff} = \operatorname{BCE} \bigl( P^a_\mathrm{no\_eff}, 1 \bigr)
  \qquad \text{and} \qquad
  L_\mathrm{pre} = \operatorname{BCE} \bigl( \bigl( \sigma(w^a) \sigma(w^a)^\top \bigr) \odot \bigl( 1 - P^a_\mathrm{no\_pre} \bigr), 1 \bigr)
  \text{.}
\end{equation}

Intuitively, minimizing $L_\mathrm{pre}$ maximizes preconditions, but the respective slots must be active for the maximization to count. Without this weighting, it would be beneficial to deactivate a slot and add arbitrary preconditions instead of keeping the slot active and learning the actual preconditions for its selection.

Notice that we use sum aggregation for the \ac{bce} loss terms above. We do not want the importance of individual elements across loss terms to depend on the relative sizes of $\hat{A}'$, $P^a_\mathrm{no\_eff}$, and $P^a_\mathrm{no\_pre}$. To this end, we introduce a common scaling factor $\frac{1}{N}$ for all loss terms, where $N$ is the total number of elements in $\hat{A}'$, $P^a_\mathrm{no\_eff}$, and $P^a_\mathrm{no\_pre}$. Excluding off-diagonal entries for unary predicates in $P^a_\mathrm{no\_eff}$ and $P^a_\mathrm{no\_pre}$, $N$ is given by
\begin{equation}
  N = R \cdot O^2 + 2 \cdot \Biggl( \sum_{r=1}^{R} M^{\abs{p_r}} \Biggr)
  \text{.}
\end{equation}

Here, $\abs{p_r}$ is the arity of predicate $p_r$. Including this scaling factor gives rise to our main loss term $L_\mathrm{main}$ and an auxiliary loss term $L_\mathrm{aux}$ weighted with $\alpha$ as
\begin{equation}
  L_\mathrm{main} = \frac{1}{N} L_\mathrm{adj}
  \qquad \text{and} \qquad
  L_\mathrm{aux} = \frac{\alpha}{N} (L_\mathrm{eff} + L_\mathrm{pre})
  \text{.}
\end{equation}

\paragraph{Gradient projection.}
Correct successor-state predictions should always take precedence over minimal effects and maximal preconditions. Rather than finding a balance, we optimize $L_\mathrm{aux}$ only under the condition that $L_\mathrm{main}$ is optimized. To do this, we separately compute gradients $\nabla L_\mathrm{aux}$ and $\nabla L_\mathrm{main}$ with respect to each parameter tensor and use PCGrad~\citep{yu_gradient_2020} (see \cref{app:grad-projection}).

\subsection{Derivation of action schemas}
\label{sec:schema-derivation}

After training, we can infer the complete action schema for each action name $a$ from the learned parameters $P^a_\mathrm{eff}$, $P^a_\mathrm{pre}$, and $w^a$. First, we apply a threshold to the slot activation probabilities to keep only clearly active slots, yielding the mask $\bar{w}^a = (\sigma(w^a) > \frac{1}{2})$. The number of active slots defines the learned arity of action $a$. For example, with $\bar{w}^a = \bmat{1 & 1 & 0 & 1 & 0}$, we get a 3-ary action $a(x_1, x_2, x_4)$.

We then compute $P^a_\mathrm{add}$, $P^a_\mathrm{del}$, $P^a_{\mathrm{pre}^+}$, and $P^a_{\mathrm{pre}^-}$ as in \cref{eq:effect-probabilities} above. The sets in the lifted action schema $\tuple{\mathrm{Add}(a(\vec{x})), \mathrm{Del}(a(\vec{x})), \mathrm{Pre}^+(a(\vec{x})), \mathrm{Pre}^-(a(\vec{x}))}$ can then be derived with a threshold of $\frac{1}{2}$, analogously to
\begin{equation}
  \label{eq:schema-derivation}
  \begin{split}
    \mathrm{Add}(a(\vec{x})) = \null&\set[\Big]{ p_r(x_i, x_j) \given[\Big] \bigl( \bigl( \bar{w}^a (\bar{w}^a)^\top \bigr) \odot P^a_\mathrm{add} \bigr)_{r,i,j} > \frac{1}{2}, \quad r \in \mathcal{P}_2, \enspace i,j \in 1, \dots, M } \\
    &\cup \set[\Big]{ p_r(x_i) \given[\Big] \bigl( \bigl( \bar{w}^a (\bar{w}^a)^\top \bigr) \odot P^a_\mathrm{add} \bigr)_{r,i,i} > \frac{1}{2}, \quad r \in \mathcal{P}_1, \enspace i \in 1, \dots, M }
    \text{.}
  \end{split}
\end{equation}

Following the injective binding assumption (see \cref{sec:problem-formulation}), each derived schema additionally requires its parameters to be pairwise distinct, i.e., $x_i \neq x_j$ for all distinct active slots $i$ and $j$.

\section{Experiments}
\label{sec:experiments}

In the following, we first describe the training and evaluation setup for \ac{dias}. We then report and discuss the main results for three action-observation variants and compare them to a symbolic method, analyze the robustness to noise, and discuss ablations, including a variant with state-dependent effects.

\subsection{Setup}

We evaluate on \num{14} standard IPC domains from the \textit{PDDL Generators} repository~\citep{seipp_pddl_2022a} or minor variations (see \cref{tab:main-results}). Training data is collected by a random walk on a single small problem instance using the ground-truth PDDL domain. We stop once every action name has at least $N_\mathrm{min}$ samples, or after $10^6$ steps. During training, we draw action-balanced batches of transitions and conduct \num{10} runs with different random seeds per configuration in the main experiments. Each run draws its own weight initialization, training walk, and test states. Domain properties, sample counts, and model and scheduling parameters are listed in \cref{app:model-training-details,app:data-all}. As stated in \cref{sec:problem-formulation}, we consider three cases for the action label~$a$:
\begin{enumerate*}[label=(\roman*)]
  \item full STRIPS actions,
  \item STRIPS+ actions, or
  \item action names only.
\end{enumerate*}
Full STRIPS actions do not require the argument-selection step from \cref{sec:argument-selection}. For the other two cases, we fix the number of slots $M$, i.e., the maximum action arity, to five. For STRIPS+ actions, the selection of known arguments is forced.

After training, we derive action schemas from the final parameters (see \cref{sec:schema-derivation}). Since arities and argument orderings may differ from the ground truth, a direct comparison is not meaningful. Instead, we compare the domains' behavior on three held-out problem instances larger than the training instance. On each, we sample \num{500} test states from multiple deep random walks, balanced over the action schemas applicable in the states (see \cref{app:testing-data}). For each test state $s$ and action name $a$, we compare the sets of reachable successor states $\mathcal{S}'_\mathrm{true}(s)$ and $\mathcal{S}'_\mathrm{pred}(s)$ under the ground-truth and learned domains. Aggregated over all test states, this yields true positives $\mathrm{TP}$, false positives $\mathrm{FP}$, and false negatives $\mathrm{FN}$, from which we report precision $\frac{\mathrm{TP}}{\mathrm{TP} + \mathrm{FP}}$ and recall $\frac{\mathrm{TP}}{\mathrm{TP} + \mathrm{FN}}$ as measures of soundness and completeness. A successor generated by the wrong action counts against both actions. Precision below 1 means that a planner using the learned domain may return invalid plans, and recall below 1 that it may miss existing ones. We also report $N_s$ for the number of runs achieving soundness with full precision, $N_c$ for runs achieving completeness with full recall, and $N_{sc}$ for runs that are both sound and complete.

\subsection{Main results}
\label{sec:main-results}

\Cref{tab:main-results} lists the main results on \num{14} domains. We first note that with full STRIPS action labels, e.g., \texttt{move(o\textsubscript{3},o\textsubscript{7},o\textsubscript{4})}, all domains except \emph{Satellite} are learned perfectly in every run. This demonstrates that the optimization objective is suitable when the argument selection is given.

The variants with STRIPS+ actions, e.g., \texttt{move(o\textsubscript{3},o\textsubscript{4})}, or action names only, e.g., \texttt{move}, both depend on learning the argument selection via a \ac{gnn} (see \cref{sec:argument-selection}). Nevertheless, in 9 of the 14 domains, we learn all action schemas perfectly in every training run with action names only, and in 11 with STRIPS+ actions. The exceptions are discussed below. \emph{Sokoban}, \emph{Hanoi}, and  \emph{Logistics} are still learned correctly in some runs, while only the two \emph{Satellite} domains fail. This demonstrates that \ac{dias} can learn correct lifted symbolic models via gradient descent.

\begin{table}[t]
  \small
  \centering
  \caption{Main results on IPC domains over \num{10} runs for three action observation variants. We report the average precision (Pr.) and recall (Re.), and the number of learned domains that are sound, complete, or sound and complete as $N_s$, $N_c$, and $N_{sc}$. Nine domains are learned perfectly in every setting. Counts with only some or zero successful runs are highlighted in orange or red, respectively.}
  \label{tab:main-results}
  \addtolength{\tabcolsep}{-0.26em}
  \begin{tabularx}{\textwidth}{X c S[table-format=1.2] S[table-format=1.2] S[table-format=2.0] S[table-format=2.0] S[table-format=2.0] c S[table-format=1.2] S[table-format=1.2] S[table-format=2.0] S[table-format=2.0] S[table-format=2.0] c S[table-format=1.2] S[table-format=1.2] S[table-format=2.0] S[table-format=2.0] S[table-format=2.0]}
    \toprule
    & & \multicolumn{5}{c}{Full STRIPS actions} & & \multicolumn{5}{c}{STRIPS+ actions} & & \multicolumn{5}{c}{Action names only} \\
    \cmidrule{3-7} \cmidrule{9-13} \cmidrule{15-19}
    Domain
    & & {Pr.} & {Re.} & {$N_s$} & {$N_c$} & {$N_{sc}$}
    & & {Pr.} & {Re.} & {$N_s$} & {$N_c$} & {$N_{sc}$}
    & & {Pr.} & {Re.} & {$N_s$} & {$N_c$} & {$N_{sc}$} \\
    \midrule
    Blocks-3 &  & 1.00 & 1.00 & 10 & 10 & 10 &  & 1.00 & 1.00 & 10 & 10 & 10 &  & 1.00 & 1.00 & 10 & 10 & 10 \\
    Delivery &  & 1.00 & 1.00 & 10 & 10 & 10 &  & 1.00 & 1.00 & 10 & 10 & 10 &  & 1.00 & 1.00 & 10 & 10 & 10 \\
    Driverlog &  & 1.00 & 1.00 & 10 & 10 & 10 &  & 1.00 & 1.00 & 10 & 10 & 10 &  & 1.00 & 1.00 & 10 & 10 & 10 \\
    Gripper &  & 1.00 & 1.00 & 10 & 10 & 10 &  & 1.00 & 1.00 & 10 & 10 & 10 &  & 1.00 & 1.00 & 10 & 10 & 10 \\
    Hanoi &  & 1.00 & 1.00 & 10 & 10 & 10 &  & 1.00 & 1.00 & 10 & 10 & 10 &  & 0.90 & 1.00 & {\cellcolor{orange!15}9} & 10 & {\cellcolor{orange!15}9} \\
    Logistics &  & 1.00 & 1.00 & 10 & 10 & 10 &  & 1.00 & 1.00 & 10 & 10 & 10 &  & 1.00 & 0.99 & 10 & {\cellcolor{orange!15}9} & {\cellcolor{orange!15}9} \\
    Miconic &  & 1.00 & 1.00 & 10 & 10 & 10 &  & 1.00 & 1.00 & 10 & 10 & 10 &  & 1.00 & 1.00 & 10 & 10 & 10 \\
    N-Puzzle &  & 1.00 & 1.00 & 10 & 10 & 10 &  & 1.00 & 1.00 & 10 & 10 & 10 &  & 1.00 & 1.00 & 10 & 10 & 10 \\
    Satellite &  & 1.00 & 0.99 & 10 & {\cellcolor{red!15}0} & {\cellcolor{red!15}0} &  & 0.93 & 0.99 & {\cellcolor{red!15}0} & {\cellcolor{red!15}0} & {\cellcolor{red!15}0} &  & 0.98 & 0.99 & {\cellcolor{red!15}0} & {\cellcolor{red!15}0} & {\cellcolor{red!15}0} \\
    Satellite-recal &  & 1.00 & 1.00 & 10 & 10 & 10 &  & 0.98 & 1.00 & {\cellcolor{red!15}0} & 10 & {\cellcolor{red!15}0} &  & 0.93 & 1.00 & {\cellcolor{red!15}0} & 10 & {\cellcolor{red!15}0} \\
    Sokoban &  & 1.00 & 1.00 & 10 & 10 & 10 &  & 1.00 & 0.98 & 10 & {\cellcolor{orange!15}2} & {\cellcolor{orange!15}2} &  & 1.00 & 0.98 & 10 & {\cellcolor{orange!15}2} & {\cellcolor{orange!15}2} \\
    Sokoban-pull &  & 1.00 & 1.00 & 10 & 10 & 10 &  & 1.00 & 1.00 & 10 & 10 & 10 &  & 1.00 & 1.00 & 10 & 10 & 10 \\
    Spanner &  & 1.00 & 1.00 & 10 & 10 & 10 &  & 1.00 & 1.00 & 10 & 10 & 10 &  & 1.00 & 1.00 & 10 & 10 & 10 \\
    Visitall &  & 1.00 & 1.00 & 10 & 10 & 10 &  & 1.00 & 1.00 & 10 & 10 & 10 &  & 1.00 & 1.00 & 10 & 10 & 10 \\
    \bottomrule
  \end{tabularx}
\end{table}

\paragraph{Analyzing failures.}
In \emph{Sokoban}, the 8 out of 10 imperfect runs with STRIPS+ actions or names only are due to dead ends, much as in symbolic methods \cite{sift}. Dead ends such as pushing a box into a corner not only prevent reaching the goal, but also block entire regions of the state space. Such domains are problematic for learning with random walks because the resulting sampling is not informative enough. Indeed, \emph{Sokoban-pull} extends \emph{Sokoban} with an extra action schema that undoes pushes, and this domain, although more complex because it contains one additional action schema, is learned perfectly. \emph{Hanoi} is still learned perfectly when some action arguments are given, but it fails in 1 of the 10 runs in the action-names-only setting. The likely reason is the same: the state space of \emph{Hanoi} resembles a Sierpiński triangle and is poorly covered by a random walk (see \cref{app:hanoi-state-space}). \emph{Logistics} is learned perfectly in every run with STRIPS+ labels and in 9 of the 10 runs with action names. In the remaining run, the learned \texttt{unload} requires a second non-airport location in the city, which only the training instance always has. We accordingly attribute this failure to the limited coverage of the training data as well. \emph{Satellite} is the only domain not learned exactly even with full STRIPS labels: all runs are sound, but none is complete. Its actions \texttt{calibrate} and \texttt{take\_image} are not well-formed STRIPS actions, since they can be applied when their add effects already hold. This also holds for \emph{Visitall}, but only in \emph{Satellite} do such transitions change nothing, as in \qty{78}{\%} of the \texttt{take\_image} and \qty{11}{\%} of the \texttt{calibrate} transitions of the random walks. Due to the gating by $p_\mathrm{pre}$, nothing refutes a precondition that only such transitions violate, such as $\neg\mathtt{calibrated}(i)$ for \texttt{calibrate}, and $L_\mathrm{pre}$ adds it. In \emph{Satellite-recal} (short for \emph{recalibration}), taking an image resets the calibration, and \texttt{calibrate} requires an uncalibrated instrument, so that every transition changes the state. With full STRIPS labels, \ac{dias} then learns the domain in every run. With the other labels, every run is complete, but none is sound due to missing preconditions on effect-less variables.

\paragraph{Learned schemas.}
When learning with action names only, the learned schemas have as many effects as the original ones in 469 of the 470 action schemas over all runs, and a matching arity in 301, with extra arguments mostly in \emph{Blocks-3}, \emph{Logistics}, and the \emph{Sokoban} domains (see \cref{app:schema-structure}). Their positive fluent preconditions are close to the original schemas (83 against 88 per run over all 47 schemas), but they add many negative fluent literals (915 against 8) and literals over static relations (1264 against 79). This is expected, since $L_\mathrm{pre}$ pushes for maximal preconditions. Since all training transitions are applicable, nothing distinguishes required from merely entailed literals. Note that with full recall, these extra preconditions do not affect correctness.

\paragraph{Comparison to a symbolic method.}
The only other approach that can learn STRIPS action schemas from traces consisting of full states and action names only, without even knowing the action arities, is the L1 algorithm by \Citet{balyo:2024}, which is based on multiple SAT calls. While the code for L1 is not publicly available, its authors ran it on one of our training walks per domain for 13 domains (all except \emph{Satellite-recal}), and we evaluate the returned domains as ours (see \cref{app:l1-results}). It thus learns from a single trace, whereas \ac{dias} is evaluated over 10 runs. L1 recovers perfect models for 6 of the 13 domains: \emph{Gripper}, \emph{Hanoi}, \emph{Miconic}, \emph{Visitall}, \emph{Delivery}, and \emph{N-Puzzle}. On the other 7, it misses preconditions that it cannot express. As confirmed by the authors, L1 cannot account for precondition variables that do not appear in action effects, as in \emph{Driverlog}, \emph{Logistics}, \emph{Satellite}, and \emph{Spanner}. It also writes negative preconditions only for atoms that the action adds and does not require distinct arguments, likely explaining the failures in \emph{Blocks-3} and both \emph{Sokoban} domains.

\subsection{Robustness to noise}
\label{sec:noise-robustness}

\Cref{tab:noise-results} shows the performance of \ac{dias} in the presence of observation noise. We consider four noise levels, where on average between one and eight randomly chosen atoms flip their observed truth value in each state transition. The flips are spread over all relations, static ones included, and only training states are corrupted. Such random flips are only a proxy for perception errors, but some tolerance to noise is required if fully symbolic states are to be replaced by other observations, such as images. During training, we slightly reduce the maximum certainty of the predicted $\hat{A}'$, which limits the influence of single corrupted atoms (see \cref{app:noise-setup}). Robustness varies strongly across domains. \emph{Delivery} and \emph{N-Puzzle} are learned exactly in every run up to two flips per transition and \emph{Sokoban-pull} at one flip, while one flip already prevents exact recovery in \emph{Blocks-3}, \emph{Driverlog}, \emph{Logistics}, \emph{Spanner}, and \emph{Visitall}. At four flips, the noise reaches or exceeds the average number of true effects per transition in all domains, yet four domains are still learned exactly in some runs, \emph{N-Puzzle} even once at eight flips. The counts $N_s$ and $N_c$ separate two failure modes. In \emph{Gripper}, \emph{Miconic}, and \emph{Visitall}, every run is sound up to four flips, and the failing runs are incomplete. In \emph{Hanoi}, \emph{Logistics}, \emph{Sokoban-pull}, and \emph{Spanner}, no failing run is sound up to four flips, while most are complete, likely because preconditions that the noise violates in a few percent of the transitions of an action are dropped. An investigation on five domains shows that more training data helps only \emph{Delivery} at two flips, and \emph{Blocks-3} and \emph{Logistics} are not learned exactly at any size (see \cref{app:learning-curves}).

\begin{table*}[t]
  \centering
  \begin{minipage}[t]{0.61\textwidth}
    \footnotesize
    \centering
    \captionof{table}{Results under observation noise over five runs with action names only, for one to eight randomly flipped atoms per transition in expectation. $N_s$, $N_c$, and $N_{sc}$ are the numbers of runs that are sound, complete, or both, with perfect and nonzero scores in green and orange, respectively. \textit{Italic} or --: only some or zero runs or test instances could be scored.}
    \label{tab:noise-results}
    \addtolength{\tabcolsep}{-0.55em}
    \begin{tabularx}{\textwidth}{X c ccc c ccc c ccc c ccc}
      \toprule
      & & \multicolumn{3}{c}{1 flip} & & \multicolumn{3}{c}{2 flips} & & \multicolumn{3}{c}{4 flips} & & \multicolumn{3}{c}{8 flips} \\
      \cmidrule{3-5} \cmidrule{7-9} \cmidrule{11-13} \cmidrule{15-17}
      Domain &  & $N_s$ & $N_c$ & $N_{sc}$ &  & $N_s$ & $N_c$ & $N_{sc}$ &  & $N_s$ & $N_c$ & $N_{sc}$ &  & $N_s$ & $N_c$ & $N_{sc}$ \\
      \midrule
        Blocks-3 &  & {\cellcolor{orange!15}1} & {\cellcolor{orange!15}4} & 0 &  & {\cellcolor{green!15}5} & 0 & 0 &  & {\cellcolor{orange!15}1} & 0 & 0 &  & 0 & 0 & 0 \\
        Delivery &  & {\cellcolor{green!15}5} & {\cellcolor{green!15}5} & {\cellcolor{green!15}5} &  & {\cellcolor{green!15}5} & {\cellcolor{green!15}5} & {\cellcolor{green!15}5} &  & {\cellcolor{orange!15}1} & {\cellcolor{green!15}5} & {\cellcolor{orange!15}1} &  & 0 & {\cellcolor{orange!15}4} & 0 \\
        Driverlog &  & 0 & {\cellcolor{orange!15}3} & 0 &  & \textit{0} & \textit{0} & \textit{0} &  & \textit{0} & \textit{0} & \textit{0} &  & \textit{0} & {\cellcolor{orange!15}\textit{2}} & \textit{0} \\
        Gripper &  & {\cellcolor{green!15}5} & {\cellcolor{orange!15}4} & {\cellcolor{orange!15}4} &  & {\cellcolor{green!15}5} & {\cellcolor{orange!15}1} & {\cellcolor{orange!15}1} &  & {\cellcolor{green!15}5} & 0 & 0 &  & 0 & 0 & 0 \\
        Hanoi &  & {\cellcolor{orange!15}3} & {\cellcolor{green!15}5} & {\cellcolor{orange!15}3} &  & 0 & {\cellcolor{green!15}5} & 0 &  & 0 & {\cellcolor{green!15}5} & 0 &  & 0 & {\cellcolor{orange!15}1} & 0 \\
        Logistics &  & 0 & {\cellcolor{green!15}5} & 0 &  & 0 & {\cellcolor{green!15}5} & 0 &  & 0 & {\cellcolor{green!15}5} & 0 &  & {\cellcolor{orange!15}1} & {\cellcolor{orange!15}1} & 0 \\
        Miconic &  & {\cellcolor{green!15}5} & {\cellcolor{orange!15}4} & {\cellcolor{orange!15}4} &  & {\cellcolor{green!15}5} & {\cellcolor{orange!15}2} & {\cellcolor{orange!15}2} &  & {\cellcolor{green!15}5} & {\cellcolor{orange!15}1} & {\cellcolor{orange!15}1} &  & 0 & 0 & 0 \\
        N-Puzzle &  & {\cellcolor{green!15}5} & {\cellcolor{green!15}5} & {\cellcolor{green!15}5} &  & {\cellcolor{green!15}5} & {\cellcolor{green!15}5} & {\cellcolor{green!15}5} &  & {\cellcolor{green!15}5} & {\cellcolor{orange!15}2} & {\cellcolor{orange!15}2} &  & {\cellcolor{orange!15}4} & {\cellcolor{orange!15}1} & {\cellcolor{orange!15}1} \\
        Satellite &  & {\cellcolor{orange!15}3} & 0 & 0 &  & \textit{0} & \textit{0} & \textit{0} &  & -- & -- & -- &  & -- & -- & -- \\
        Satellite-recal &  & 0 & 0 & 0 &  & \textit{0} & \textit{0} & \textit{0} &  & -- & -- & -- &  & -- & -- & -- \\
        Sokoban &  & {\cellcolor{orange!15}3} & {\cellcolor{orange!15}2} & {\cellcolor{orange!15}1} &  & {\cellcolor{orange!15}4} & {\cellcolor{orange!15}4} & {\cellcolor{orange!15}3} &  & 0 & {\cellcolor{orange!15}1} & 0 &  & \textit{0} & \textit{0} & \textit{0} \\
        Sokoban-pull &  & {\cellcolor{green!15}5} & {\cellcolor{green!15}5} & {\cellcolor{green!15}5} &  & {\cellcolor{orange!15}3} & {\cellcolor{green!15}5} & {\cellcolor{orange!15}3} &  & {\cellcolor{orange!15}3} & {\cellcolor{green!15}5} & {\cellcolor{orange!15}3} &  & 0 & {\cellcolor{green!15}5} & 0 \\
        Spanner &  & 0 & {\cellcolor{green!15}5} & 0 &  & 0 & {\cellcolor{green!15}5} & 0 &  & 0 & {\cellcolor{orange!15}3} & 0 &  & 0 & {\cellcolor{orange!15}3} & 0 \\
        Visitall &  & {\cellcolor{green!15}5} & 0 & 0 &  & {\cellcolor{green!15}5} & 0 & 0 &  & {\cellcolor{green!15}5} & 0 & 0 &  & {\cellcolor{orange!15}4} & 0 & 0 \\
      \bottomrule
    \end{tabularx}
  \end{minipage}
  \hfill
  \begin{minipage}[t]{0.36\textwidth}
    \footnotesize
    \centering
    \captionof{table}{Results of the \ac{mlp}-effects variant with action names only over \num{10} runs. We report average precision (Pr.), recall (Re.), and $N_s$, $N_c$, and $N_{sc}$ as before. Counts with only some or zero successful runs are highlighted in orange or red.\vspace{0.35em}}
    \label{tab:mlp-baseline-results}
    \addtolength{\tabcolsep}{-0.45em}
    \begin{tabularx}{\textwidth}{X c S[table-format=1.2] S[table-format=1.2] S[table-format=2.0] S[table-format=2.0] S[table-format=2.0]}
      \toprule
      {Domain} & & {Pr.} & {Re.} & {$N_s$} & {$N_c$} & {$N_{sc}$} \\
      \midrule
        Blocks-3 &  & 1.00 & 1.00 & 10 & 10 & 10 \\
        Delivery &  & 1.00 & 1.00 & 10 & 10 & 10 \\
        Driverlog &  & 0.82 & 0.99 & {\cellcolor{orange!15}8} & {\cellcolor{red!15}0} & {\cellcolor{red!15}0} \\
        Gripper &  & 1.00 & 1.00 & 10 & 10 & 10 \\
        Hanoi &  & 0.92 & 0.99 & {\cellcolor{orange!15}9} & {\cellcolor{orange!15}9} & {\cellcolor{orange!15}8} \\
        Logistics &  & 0.40 & 1.00 & {\cellcolor{red!15}0} & 10 & {\cellcolor{red!15}0} \\
        Miconic &  & 1.00 & 1.00 & 10 & 10 & 10 \\
        N-Puzzle &  & 1.00 & 1.00 & 10 & 10 & 10 \\
        Satellite &  & 0.95 & 0.99 & {\cellcolor{red!15}0} & {\cellcolor{red!15}0} & {\cellcolor{red!15}0} \\
        Satellite-recal &  & 0.95 & 1.00 & {\cellcolor{red!15}0} & 10 & {\cellcolor{red!15}0} \\
        Sokoban &  & 1.00 & 0.99 & 10 & {\cellcolor{orange!15}5} & {\cellcolor{orange!15}5} \\
        Sokoban-pull &  & 1.00 & 1.00 & 10 & 10 & 10 \\
        Spanner &  & 0.21 & 1.00 & {\cellcolor{red!15}0} & 10 & {\cellcolor{red!15}0} \\
        Visitall &  & 1.00 & 0.97 & 10 & {\cellcolor{orange!15}9} & {\cellcolor{orange!15}9} \\
      \bottomrule
    \end{tabularx}
  \end{minipage}
\end{table*}

\subsection{Ablations}
\label{sec:ablations}

We study ablations of our approach and find that removing any component degrades the results (see \cref{app:component-ablations}). Of 70 runs with action names, \ac{dias} learns 56 exactly, 43 without $L_\mathrm{eff}$, 30 without slot weights, 12 without $L_\mathrm{pre}$, 7 without both auxiliary terms, and 5 without PCGrad. Without $L_\mathrm{pre}$, 64 runs remain complete, but only those 12 are sound. The results change little for derivation thresholds from 0.25 to 0.75, for $\tau$ schedules five times faster or twice as slow, and for 7 instead of 5 slots (52 to 56 runs, \cref{app:sensitivity}). A single $\alpha$ for all domains, however, recovers at most 49 runs.

\paragraph{MLP effects.}
To test whether learning depends on the STRIPS form of the effects, we replace only the lifted effect parameters $P^a_\mathrm{eff}$ with an \ac{mlp} that predicts the effects from the current state, and keep the argument selection and the preconditions. The input of the \ac{mlp} is the lifted state $\tilde{S}^a A (\tilde{S}^a)^\top$, i.e., the truth values in $s$ of all atoms over the objects selected into the slots, and its output has the shape of $P^a_\mathrm{eff}$ (see \cref{app:mlp-effects}). The result is a slot-based dynamics model in the spirit of \acp{rim} \cite{rim}. It can express state-dependent effects, which would require conditional effects in PDDL. Unlike \acp{rim}, it generalizes to larger instances even in domains such as \emph{Blocks}, where more objects mean more state variables with more possible values. Its effects, however, are not a PDDL schema, so a planner would have to query the network for every successor. With the hyperparameters of \ac{dias} and action names only, it learns 82 of 140 runs exactly, against 110 for \ac{dias}, as \cref{tab:mlp-baseline-results} shows. Of the nine domains that \ac{dias} learns in every run, it matches on six. It fails to learn \emph{Driverlog}, \emph{Logistics}, or \emph{Spanner} at all, with the latter two being complete but not sound. It is better than \ac{dias} only in \emph{Sokoban} (5 against 2 runs). It thus opens up interesting possibilities and shows that the STRIPS form of the effects is not required to learn the argument selection but makes learning more reliable, and only STRIPS effects yield schemas that a planner can use efficiently.

\section{Conclusions}
\label{sec:conclusions}

We have developed a novel neural architecture for learning STRIPS action schemas from traces in which states are fully observed but action arguments are hidden. The main challenge is to learn the action schemas while simultaneously identifying the action arguments from observed state changes, especially when preconditions involve variables that do not appear in the action effects. We evaluated the proposed architecture, \ac{dias}, on 14 classical planning domains and found that, from action names alone, it learns the ground-truth schemas of 9 domains in every run and in 3 more domains in some runs. We also analyzed the failure cases and robustness to noise, and compared \ac{dias} to both a symbolic baseline and a slot-based dynamic-model variant that uses an \ac{mlp} instead of STRIPS effects. In future work, we aim to use \ac{dias} as a differentiable component for learning action schemas reliably from transitions of state images instead of symbolic states.

\paragraph{Limitations.}
\ac{dias} relies on the assumptions of \cref{sec:problem-formulation}: fully observed states over known unary and binary predicates, injective binding of action arguments, and \strips dynamics. In practice, the results depend on data coverage and, in some domains, on the auxiliary-loss weight $\alpha$. Preconditions are also harder to identify if an action has no effect in some states or if they involve arguments that do not appear in the effects. The learned schemas are not guaranteed to be equivalent to the originals, and their preconditions may be redundant. Noise robustness remains limited, even with more data, and missing or noisy action names are not handled. The future work of learning from images would additionally require a perception module that maps images to symbolic states.

\section*{Acknowledgments}

We thank Jonas Gösgens, Michael Aichmüller, and especially Niklas Jansen for insightful comments and valuable discussions. Furthermore, we are grateful to Tomáš Balyo for his cooperation in providing the domains that his method learned from our traces. The research has been supported by the Alexander von Humboldt Foundation with funds from the German Federal Ministry for Education and Research. This project has received funding from the European Research Council (ERC) under the European Union’s Horizon 2020 research and innovation programme (Grant agreement No. 885107). This project was also funded by the German Federal Ministry of Education and Research (BMBF) and the Ministry of Culture and Science of the German State of North Rhine-Westphalia (MKW) under the Excellence Strategy of the Federal Government and the Länder.

\printbibliography

\clearpage
\appendix

\section{Related work}
\label{app:related-work}

The domain learning problem considered here has been studied across symbolic and neural lines of work, under different assumptions about observability, action labels, noise, and the representation of states. We discuss four related research threads and position our approach relative to them.

\paragraph{Symbolic model learning in classical planning.}
The problem of learning lifted \strips models from state-action traces has received considerable attention \cite{zhuo2013action,aineto2019learning,lamanna2021online,verma2021asking,le2024learning,bachor2024learning,aineto2024action,paolo:2025}. While observability of the states can be partial or noisy, in almost all cases the observations are assumed to reveal all predicates and all action arguments. Exceptions include the SAT-based approaches of \Citet{bonet:ecai2020} and \Citet{ivan:kr2021}, where only action names and state equalities need to be observed (i.e., whether two states are the same without knowing their contents). More recently, \sift learns lifted models from action traces alone, assuming that the actions are full \strips actions, as in the earlier LOCM systems \cite{locm1,Locm}. Domain predicates are learned along the way, and the algorithm is provably correct and efficient. The limitation is that assuming all \strips action arguments are observed is often unrealistic. The recent \synth algorithm \cite{jansen_learning_2025} instead assumes that states and non-redundant action arguments are observable, while \Citet{balyo:2024} introduce a SAT-based formulation where only states and action names are observed, as in our setting.

\paragraph{Neural model learning in classical planning.}
Works addressing \strips model learning via gradient descent include LatPlan \cite{asai:latplan,asai:jair}, a transformer-based architecture \cite{nunez_molina_next_2025_jakob}, and the ROSAME~1 and~2 systems \cite{xi_neurosymbolic_2024,xi_learning_2026}. The first two learn propositional \strips models, while ROSAME~1 learns lifted \strips models from traces made up of images and full \strips actions, and ROSAME~2 learns from traces made up of images and action names only. The latter is closest to our work, although it addresses the more challenging setting in which only the initial and final symbolic states of the traces are given, while the intermediate symbolic states are replaced by images. The accuracy of the learned \strips models, however, remains limited. We aim instead at the simpler problem where all symbolic states in the traces are given, possibly with noise, with the goal of solving it nearly perfectly.

\paragraph{Model-based reinforcement learning.}
Model-based reinforcement learning algorithms learn controllers by also learning (stochastic) \ac{mdp} models, without making assumptions about their internal structure \cite{sutton:book}. In the tabular setting, they result in flat state models with transition probabilities obtained from simple counts \cite{brafman:rmax}. In some cases, a first-order state language is assumed, but the state predicates are given \cite{littman:rmax,leslie:probabilistic}. In more recent approaches, the learned dynamics is represented not in compact languages such as \strips or probabilistic PDDL \cite{littman:competition}, but in terms of deep neural networks \cite{fleuret:atari,dreamer:atari,timofte:atari}. These models are useful for control, but they are not lifted and thus do not generalize naturally to new objects.

\paragraph{Slot-based dynamic model learning.}
Slot-based dynamic models in deep learning, such as \acp{rim} \cite{rim} and related approaches, reduce states to the values of a fixed number of slots that play the role of state variables. It is further assumed that only a few slots change value at any time point, and that these changes are a function of their previous values. These models define an implicit representation language over state variables, but the main evaluation focus is usually predictive performance rather than whether the intended symbolic structure is actually recovered. Our architecture differs in that it does not select a subset of state variables at each point, but a subset of objects, which in turn define a fixed set of atoms or Boolean variables. This distinction matters for generalization to more objects, because then both the number of state variables and the number of values they can take may change. This limitation does not arise in our setting, where the learned lifted representations generalize to an arbitrary number of objects.

\section{Model and training}
\label{app:model-training-details}

This section follows one transition through the model, lists the architecture and optimization settings, and details the gradient projection and the Sinkhorn normalization.

\subsection{Walkthrough: one \emph{Blocks-3} transition}
\label{app:walkthrough}

Consider a transition in which $\mathtt{move}$ moves block $b_1$ from $b_2$ onto $b_3$. The state $s$ contains $\mathtt{on}(b_1,b_2)$, $\mathtt{clear}(b_1)$, $\mathtt{clear}(b_3)$, $\mathtt{on\text{-}table}(b_2)$, and $\mathtt{on\text{-}table}(b_3)$, among others. In $s'$, $\mathtt{on}(b_1,b_3)$ and $\mathtt{clear}(b_2)$ are added, and $\mathtt{on}(b_1,b_2)$ and $\mathtt{clear}(b_3)$ are deleted.

\paragraph{Graph.}
The objects are the nodes. $\mathtt{on}(b_1,b_2)$ becomes an edge of type $\mathtt{on}$ from $b_1$ to $b_2$, the unary atoms become self-loops, and the four changes become edges of the types $\mathtt{on}'_\mathrm{add}$ (from $b_1$ to $b_3$), $\mathtt{clear}'_\mathrm{add}$, $\mathtt{on}'_\mathrm{del}$, and $\mathtt{clear}'_\mathrm{del}$. Every edge also gets a reverse edge. With action names only, there are no argument self-loops.

\paragraph{Selection.}
The \ac{rgcn} computes the keys $K$, which are matched against the queries $Q^\mathtt{move}$ of the five slots, numbered 0 to 4 as in \cref{lst:blocks-3-diff}. Suppose that Sinkhorn assigns $b_1$ to slot 1, $b_3$ to slot 2, and $b_2$ to slot 3, and that $\sigma(w^\mathtt{move}_0)$ and $\sigma(w^\mathtt{move}_4)$ are close to zero, so that $\tilde{S}^\mathtt{move}$ suppresses slots 0 and 4, whichever objects they select.

\paragraph{Prediction.}
The projection maps each lifted atom over slots to the ground atom over the objects in these slots. The add effect $\mathtt{on}(x_1, x_2)$ thus becomes $\mathtt{on}(b_1, b_3)$, and likewise for the other effects. The preconditions $\mathtt{clear}(x_1)$, $\mathtt{clear}(x_2)$, and $\mathtt{on}(x_1, x_3)$ hold in $s$ for the selected objects, so $p_\mathrm{pre} \approx 1$, and the prediction $\hat{A}'$ matches $A'$.

\paragraph{Extraction.}
The mask $\bar{w}^\mathtt{move} = (0, 1, 1, 1, 0)$ gives the schema $\mathtt{move}(x_1, x_2, x_3)$. Its add effects are $\mathtt{on}(x_1, x_2)$ and $\mathtt{clear}(x_3)$, its delete effects are $\mathtt{on}(x_1, x_3)$ and $\mathtt{clear}(x_2)$, and its preconditions include $\mathtt{clear}(x_1)$, $\mathtt{clear}(x_2)$, $\mathtt{on}(x_1, x_3)$, and the pairwise distinctness of $x_1$, $x_2$, and $x_3$, as in \cref{lst:blocks-3-diff}.

\subsection{Architecture and optimization}
\label{app:model-configuration}

All runs share the following configuration unless stated otherwise.

\paragraph{Encoder.}
The \ac{rgcn} has separate edge types for the atoms of each relation in $s$, for its added atoms, and for its deleted atoms, each also in reverse direction, and one for each observed argument position of each action, i.e., $6R$ edge types with action names only. It has six layers with sum aggregation and output widths 32, 64, 64, 64, 64, and $d = 32$. Each layer but the last is followed by SiLU and layer normalization, and the last has no bias. The initial node features are redrawn in every forward pass: the first $d/2$ entries are drawn from $0.1 \cdot \mathcal{N}(0, 1)$ and the others are zero, so that the noise breaks the symmetry between objects, while the zero half keeps a common anchor.

\paragraph{Optimization.}
We train for \num{10000} steps with AdamW~\citep{loshchilov:decoupled_iclr}, a learning rate of \num{5e-3}, and action-balanced batches of \num{200} transitions. The number of slots is $M = 5$, except with full STRIPS labels, where each action has one slot per argument. The root degree in \cref{eq:pre-factor-schedule} decays as $\tau = 0.1^{t/500}$ in step $t$, i.e., from the geometric mean at $\tau = 1$ to $\tau = 0.1$ at step 500 and further toward the plain product. Training is deterministic given the domain and the seed.

\subsection{Gradient projection}
\label{app:grad-projection}

For each parameter tensor, let $g_\mathrm{main}$ and $g_\mathrm{aux}$ be the gradients of $L_\mathrm{main}$ and $L_\mathrm{aux}$, where $g_\mathrm{aux}$ includes the weight $\alpha$. Where the two oppose each other, PCGrad~\citep{yu_gradient_2020} removes the component of $g_\mathrm{aux}$ along $g_\mathrm{main}$,
\begin{equation}
  \tilde{g}_\mathrm{aux} =
  \begin{cases}
    g_\mathrm{aux} - \frac{g_\mathrm{aux} \cdot g_\mathrm{main}}{\norm{g_\mathrm{main}}^2} \, g_\mathrm{main} & \text{if} \enspace g_\mathrm{aux} \cdot g_\mathrm{main} < 0 \text{,} \\
    g_\mathrm{aux} & \text{otherwise,}
  \end{cases}
\end{equation}
and we cap its norm by that of the main gradient,
\begin{equation}
  g = g_\mathrm{main} + \min \Bigl( 1, \frac{\norm{g_\mathrm{main}}}{\norm{\tilde{g}_\mathrm{aux}}} \Bigr) \, \tilde{g}_\mathrm{aux}
  \text{.}
\end{equation}
The projection is skipped if $\norm{g_\mathrm{main}}^2 \le 10^{-12}$, and a tensor on which $L_\mathrm{main}$ does not depend is updated with $g_\mathrm{aux}$ alone. The cap matters near a minimum of $L_\mathrm{main}$, where $g_\mathrm{main}$ becomes small and noisy and an uncapped $\tilde{g}_\mathrm{aux}$ would move the parameters away from it. As a consequence, $\alpha$ only takes effect while $\norm{\tilde{g}_\mathrm{aux}} < \norm{g_\mathrm{main}}$. \Cref{app:sensitivity} discusses the sensitivity to $\alpha$, and \cref{tab:domain-config} lists the values we use.

\subsection{Rectangular Sinkhorn normalization}
\label{app:sinkhorn}

The correspondence matrix $C^a \in \mathbb{R}^{M \times O}$ is rectangular, since the number of slots $M$ is fixed while the number of objects $O$ varies, so the doubly stochastic target of the standard Sinkhorn algorithm does not apply. \Citet{brun_differentiable_2022} add a slack row and a slack column for unmatched elements. We add only the slack row, initialized to zero and exempt from row normalization, so that it absorbs the mass of objects that no slot selects, and discard it after convergence. The resulting $S^a$ has row sums of 1 and column sums of at most 1. Slots that select no object are instead handled by the slot weights $\sigma(w^a)$ of \cref{sec:argument-selection}. To avoid underflow, we iterate in log-space on the extended matrix $\hat{C}^a \in \mathbb{R}^{(M+1) \times O}$, with row and column potentials $u$ and $v$ initialized to zero,
\begin{equation*}
  u_k \leftarrow -\operatorname{logsumexp}_i \bigl(\hat{C}^a_{k,i} + v_i\bigr) \enspace (k \le M), \quad u_{M+1} \leftarrow 0, \quad v_i \leftarrow -\operatorname{logsumexp}_k \bigl(\hat{C}^a_{k,i} + u_k\bigr),
\end{equation*}
until no entry of $u$ and $v$ changes by more than $10^{-2}$ in one iteration, or for at most \num{10000} iterations, and obtain $S^a_{k,i} = \exp(u_k + \hat{C}^a_{k,i} + v_i)$ for $k \le M$. All Sinkhorn computations run in \texttt{float32}, even under mixed precision.

\section{Experimental setup}
\label{app:data-all}

This section describes the domains, how we collect the training and test data and score the learned domains, the setups of the noise experiments and the \ac{mlp}-effects variant, and the compute we used.

\subsection{Domains}
\label{app:per-domain}

\Cref{tab:domain-signatures} lists the actions and predicates of the original PDDL domains with their arities, and \cref{tab:domain-config} the numbers of objects in the training and test instances, the per-action sample bounds, and the auxiliary-loss weight $\alpha$ of each of the \num{14} domains. \emph{Satellite-recal} differs from \emph{Satellite} in that \texttt{take\_image} deletes \texttt{calibrated}, and \texttt{calibrate} requires the instrument not to be calibrated.

% |\arity{k}| typesets the arity k as a gray subscript in \cref{tab:domain-signatures}.
\newcommand{\arity}[1]{\raisebox{-0.4ex}{\fontsize{7}{7}\selectfont\textcolor{black!55}{#1}}}
\begin{table}[!ht]
  \centering
  \small
  \caption{Actions and predicates of the original PDDL domains, with their arities as gray subscripts.}
  \label{tab:domain-signatures}
  \begin{tabularx}{\linewidth}{@{}l >{\raggedright\arraybackslash\ttfamily}X >{\raggedright\arraybackslash\ttfamily}X@{}}
    \toprule
    Domain & \normalfont Actions & \normalfont Predicates \\
    \midrule
    Blocks-3 & stack\arity{2}, newtower\arity{2}, move\arity{3} & clear\arity{1}, on-table\arity{1}, on\arity{2} \\\addlinespace
    Delivery & pick-package\arity{3}, drop-package\arity{3}, move\arity{3} & at\arity{2}, carrying\arity{2}, empty\arity{1}, adjacent\arity{2} \\\addlinespace
    Driverlog & load-truck\arity{3}, unload-truck\arity{3}, board-truck\arity{3}, disembark-truck\arity{3}, drive-truck\arity{4}, walk\arity{3} & at\arity{2}, in\arity{2}, driving\arity{2}, link\arity{2}, path\arity{2}, empty\arity{1} \\\addlinespace
    Gripper & move\arity{2}, pick\arity{3}, drop\arity{3} & room\arity{1}, ball\arity{1}, gripper\arity{1}, at-robby\arity{1}, at\arity{2}, free\arity{1}, carry\arity{2} \\\addlinespace
    Hanoi & move\arity{3} & clear\arity{1}, on\arity{2}, smaller\arity{2} \\\addlinespace
    Logistics & load\arity{3}, unload\arity{3}, drive\arity{4}, fly\arity{3} & package\arity{1}, truck\arity{1}, airplane\arity{1}, vehicle\arity{1}, location\arity{1}, airport\arity{1}, city\arity{1}, loc\arity{2}, at\arity{2}, in\arity{2} \\\addlinespace
    Miconic & unboard\arity{2}, board\arity{2}, move\_up\arity{2}, move\_down\arity{2} & person\arity{1}, floor\arity{1}, lift\_pos\arity{1}, in\_lift\arity{1}, in\_floor\arity{2}, above\arity{2} \\\addlinespace
    N-Puzzle & move-up\arity{3}, move-down\arity{3}, move-left\arity{3}, move-right\arity{3} & tile\arity{1}, cell\arity{1}, at\arity{2}, blank\arity{1}, above\arity{2}, right\arity{2} \\\addlinespace
    Satellite & turn\_to\arity{3}, switch\_on\arity{2}, switch\_off\arity{2}, calibrate\arity{3}, take\_image\arity{4} & on\_board\arity{2}, supports\arity{2}, pointing\arity{2}, power\_avail\arity{1}, power\_on\arity{1}, calibrated\arity{1}, have\_image\arity{2}, calibration\_target\arity{2} \\\addlinespace
    Satellite-recal & turn\_to\arity{3}, switch\_on\arity{2}, switch\_off\arity{2}, calibrate\arity{3}, take\_image\arity{4} & on\_board\arity{2}, supports\arity{2}, pointing\arity{2}, power\_avail\arity{1}, power\_on\arity{1}, calibrated\arity{1}, have\_image\arity{2}, calibration\_target\arity{2} \\\addlinespace
    Sokoban & move-player\arity{2}, push-box\arity{3} & has\_player\arity{1}, has\_box\arity{1}, adjacent\arity{2}, adjacent\_2\arity{2} \\\addlinespace
    Sokoban-pull & move-player\arity{2}, push-box\arity{3}, pull-box\arity{3} & has\_player\arity{1}, has\_box\arity{1}, adjacent\arity{2}, adjacent\_2\arity{2} \\\addlinespace
    Spanner & walk\arity{3}, pickup\_spanner\arity{3}, tighten\_nut\arity{4} & at\arity{2}, carrying\arity{2}, useable\arity{1}, link\arity{2}, tightened\arity{1}, loose\arity{1} \\\addlinespace
    Visitall & move\arity{2} & connected\arity{2}, at-robot\arity{1}, visited\arity{1} \\
    \bottomrule
  \end{tabularx}
\end{table}

\begin{table}[!ht]
  \centering
  \small
  \caption{Numbers of objects in the training and the three test instances, per-action sample bounds of the training split, and auxiliary-loss weight $\alpha$. The bounds of the validation split are half of those of the training split.}
  \label{tab:domain-config}
  \begin{tabular}{@{}l c cccc c cc c c@{}}
    \toprule
    & & \multicolumn{4}{c}{Objects} & & \multicolumn{2}{c}{Transitions} & & \\
    \cmidrule{3-6} \cmidrule{8-9}
    Domain & & Train & Test-1 & Test-2 & Test-3 & & $N_\mathrm{min}^\mathrm{tr}$ & $N_\mathrm{max}^\mathrm{tr}$ & & $\alpha$ \\
    \midrule
    Blocks-3 & & 5 & 10 & 20 & 40 & & 100 & 1000 & & 1 \\
    Delivery & & 15 & 21 & 45 & 81 & & 1000 & 2000 & & 0.1 \\
    Driverlog & & 16 & 24 & 48 & 89 & & 2000 & 3000 & & 0.01 \\
    Gripper & & 10 & 12 & 22 & 42 & & 200 & 1000 & & 1 \\
    Hanoi & & 10 & 15 & 27 & 51 & & 1000 & 1000 & & 0.3 \\
    Logistics & & 15 & 24 & 36 & 48 & & 100 & 1000 & & 0.3 \\
    Miconic & & 10 & 10 & 20 & 40 & & 100 & 1000 & & 1 \\
    N-Puzzle & & 17 & 31 & 49 & 71 & & 100 & 1000 & & 1 \\
    Satellite & & 36 & 72 & 80 & 89 & & 100 & 1000 & & 0.01 \\
    Satellite-recal & & 36 & 72 & 80 & 89 & & 100 & 1000 & & 0.01 \\
    Sokoban & & 16 & 36 & 49 & 81 & & 100 & 1000 & & 0.1 \\
    Sokoban-pull & & 16 & 36 & 49 & 81 & & 100 & 1000 & & 0.1 \\
    Spanner & & 19 & 35 & 67 & 88 & & 200 & 1000 & & 1 \\
    Visitall & & 25 & 36 & 49 & 64 & & 100 & 1000 & & 1 \\
    \bottomrule
  \end{tabular}
\end{table}

\subsection{Training data}
\label{app:training-data}

For each domain, we sample transitions by a random walk in the training instance (\texttt{pddl/<domain>/train.pddl} in our code). Starting from the initial state, each step applies a ground action drawn uniformly from those applicable, and the walk returns to the initial state at dead ends. It stops as soon as every action name has at least $N_\mathrm{min}^\mathrm{tr} + N_\mathrm{min}^\mathrm{val}$ transitions, or after $10^6$ steps. Of each action name, we keep the first $N_\mathrm{max}^\mathrm{tr} + N_\mathrm{max}^\mathrm{val}$ transitions and split all kept transitions at random in the ratio $N_\mathrm{min}^\mathrm{tr} : N_\mathrm{min}^\mathrm{val} = 2 : 1$ into training and validation transitions, such that every action name with at least two transitions appears in both. The validation split is only used for monitoring. The bounds are listed in \cref{tab:domain-config}. As for $\alpha$ (\cref{app:sensitivity}), we chose them for each domain in preliminary experiments, based on the validation results of a few runs. Only in \emph{Sokoban} does the walk reach the step limit: after a few pushes, no box can be pushed anymore, and the walk collects only a handful of \texttt{push-box} transitions. In none of our domains does an applicable ground action bind one object to two arguments. The seed of a run determines its training walk, the model initialization, and the random node features. The configurations of all runs are part of our code.

\subsection{Test data and scoring}
\label{app:testing-data}

For every domain, we evaluate on three larger held-out instances (\texttt{test-1} to \texttt{test-3}, see \cref{tab:domain-config}). On each, we draw 500 distinct test states, an equal share for each action schema. We repeat walks of \num{2000} steps from the initial state, defined as for training, at least five and at most 100 times, until every action schema is applicable in at least $500 / \abs{\mathcal{A}}$ of the distinct visited states. For each action schema, we then draw its share uniformly from the visited states in which it is applicable and that were not drawn yet, and fill up any shortfall from the remaining visited states. We sample by walks rather than breadth-first search because the states that a breadth-first search reaches first lie within a few actions of the initial state, and on the larger instances, some actions, such as \texttt{take\_image} in \emph{Satellite}, are not applicable in any of them. The test states are drawn anew for every run with a recorded seed.

For a test state $s$ and an action name $a$, let $\mathcal{S}'_\mathrm{true}(s, a)$ and $\mathcal{S}'_\mathrm{pred}(s, a)$ be the successors under all groundings of $a$ in the original and in the learned domain. $\mathrm{TP}$ counts the successors in both sets, $\mathrm{FP}$ those only in $\mathcal{S}'_\mathrm{pred}(s, a)$, and $\mathrm{FN}$ those only in $\mathcal{S}'_\mathrm{true}(s, a)$, summed over all action names and the \num{1500} test states of the three instances. A run is sound if its precision is 1 and complete if its recall is 1. If a learned domain generates no successor at all, we set its precision to 0. We compute both successor sets by merging the learned actions into the original domain and grounding all actions in $s$. A run whose learned domain cannot be grounded within the memory limit on an instance is scored on the remaining instances, or left out if no instance remains.

\subsection{Noise experiments}
\label{app:noise-setup}

For the experiments in \cref{sec:noise-robustness}, we corrupt the training states and slightly adjust the training.

\paragraph{Noise model.}
We corrupt the states $s$ and $s'$ of every training transition independently. For a state and each relation $r$, including static relations and types, let $T_r$ and $F_r$ be the sets of its true and false ground atoms, over all $O$ objects for unary relations and all $O^2$ pairs of objects for binary ones. With $k$ flips per transition, we draw the numbers of atoms removed from and added to each relation independently as
\begin{equation*}
  n^-_r, n^+_r \sim \operatorname{Poisson}\Bigl(\frac{k}{4}\, w_r\Bigr), \qquad w_r = \frac{\min(\abs{T_r}, \abs{F_r})}{\sum_q \min(\abs{T_q}, \abs{F_q})},
\end{equation*}
and remove $\min(n^-_r, \abs{T_r})$ atoms drawn uniformly without replacement from $T_r$ and add $\min(n^+_r, \abs{F_r})$ atoms drawn in the same way from $F_r$. Summed over all relations, the numbers of removed and added atoms are again Poisson distributed with mean $k/4$ each, so a transition receives $k$ flips in expectation, apart from the rare draws that exceed the available atoms. The weights $w_r$ direct the noise to relations with both many true and many false atoms, keep it from growing with the number of objects, and leave relations whose atoms are all true or all false clean. Since static relations and types are included, most flips hit static atoms, e.g., \qty{47}{\%} in \emph{Blocks-3} and \qty{96}{\%} in \emph{Sokoban}. Training transitions are corrupted once per run, and test states are not corrupted.

\paragraph{Training adjustments.}
We reduce the batch size to \num{50} and read the prediction as observed through a channel that flips each atom with probability $\varepsilon = 0.1$, i.e., we use $L_\mathrm{adj} = \operatorname{BCE}(\varepsilon + (1 - 2\varepsilon)\hat{A}', A')$. Since the prediction then lies in $[\varepsilon, 1 - \varepsilon]$, each atom costs at most $-\log \varepsilon \approx 2.3$, and the gradient with respect to $\hat{A}'_{r,i,j}$ is at most $(1 - 2\varepsilon) / \varepsilon = 8$ in magnitude. A single corrupted atom thus cannot outweigh the transitions that agree with the prediction, and $\tau$ can decay to $0$ as without noise. All other settings are those of the noise-free experiments.

\subsection{MLP-effects variant}
\label{app:mlp-effects}

The \ac{mlp}-effects variant of \cref{sec:ablations}, whose results are listed in \cref{tab:mlp-baseline-results}, replaces the static effect parameters $P^a_\mathrm{eff}$ with a shared \ac{mlp} that predicts them from the state. The preconditions remain static parameters $P^a_\mathrm{pre}$, for the reason given at the end of this section.

\paragraph{Training.}
The keys $K$ and the selection $\tilde{S}^a$ are computed as in \ac{dias}. The \ac{mlp} reads the state restricted to the selected objects and predicts the effect logits,
\begin{equation}
  \label{eq:state-dependent-effects}
  \tilde{A}^a = \tilde{S}^a A (\tilde{S}^a)^\top \in (0, 1)^{R \times M \times M}
  \qquad \text{and} \qquad
  P^a_\mathrm{eff} = \operatorname{MLP}^a \bigl( \tilde{A}^a \bigr) \in \mathbb{R}^{R \times M \times M \times 3}
  \text{,}
\end{equation}
where $\operatorname{MLP}^a$ is the output slice for action $a$ of one \ac{mlp} shared by all actions, with two hidden layers of width 64 and SiLU activations. As for the static effects, unary relations only have effects on the diagonal. $P^a_\mathrm{add}$ and $P^a_\mathrm{del}$ follow from \cref{eq:effect-probabilities}, and $\hat{A}'$ from \cref{eq:effect-application}. The variant uses the hyperparameters of \ac{dias} (\cref{app:model-configuration}) without additional tuning.

\paragraph{Evaluation.}
At test time, $s'$ is unknown, so the selection $\tilde{S}^a$ cannot be computed. Instead, we threshold $w^a$ and $P^a_\mathrm{pre}$ as in \cref{sec:schema-derivation} and write the result as a PDDL action without effects over the active slots. Grounding it in a test state $s$ yields the applicable bindings of objects to the active slots. For each binding, we replace $\tilde{S}^a$ by the corresponding one-hot selection matrix, let the \ac{mlp} predict the effects from the atoms over the bound objects, apply them with $p_\mathrm{pre} = 1$ as in \cref{eq:effect-application}, and round $\hat{A}'$ at $\frac{1}{2}$, which gives one successor in $\mathcal{S}'_\mathrm{pred}(s, a)$. Scoring then proceeds as for \ac{dias}.

\paragraph{Static preconditions.}
\label{app:no-precondition-mlp}
Preconditions predicted from the state would admit a trivial solution, since all training transitions are applicable. An \ac{mlp} that predicts $P^a_\mathrm{pre}$ from $s$ maximizes $p_\mathrm{pre}$ in \cref{eq:effect-application} by declaring every atom of $s$ a positive precondition, which every observed transition satisfies. Learning such preconditions would require negative samples, i.e., transitions whose action is not applicable, which we leave to future work.

\subsection{Compute requirements}
\label{app:compute-requirements}

All experiments were run on our institute's compute cluster. Each training run used 8 CPU cores, between 4 and \qty{32}{\giga B} of RAM, and between 4 and \qty{12}{\giga B} of the memory of an NVIDIA A10 GPU (\qty{24}{\giga B}), so that two to six runs shared one card. The median training run took 47 minutes, \qty{90}{\%} finished within about two hours, and the longest took 22.5 hours. The learned domains were validated on CPUs, mostly with 8 cores and 8 to \qty{16}{\giga B} of RAM, and the median validation took under a minute. Only the validation of the \ac{mlp}-effects variant also used a share of a GPU, and the scaling study in \cref{tab:scaling} used one A10 per task. In total, the \num{2240} training runs and \num{2603} validation jobs behind our tables used about \num{43000} CPU core-hours and \num{1060} GPU-hours, counting each run's share of a card, or about \num{3800} GPU-hours with one card per run. Beyond these runs, the total compute consumed during the project is substantially larger due to hyperparameter tuning, architectural exploration, and failed experiments that did not make it into the paper.

\section{Additional results}
\label{app:results}

This section complements \cref{sec:experiments} with an example and statistics of the learned schemas, the coverage of the \emph{Hanoi} state space, the full results of L1, learning curves, ablations, sensitivity, and scalability.

\subsection{Learned schemas}
\label{app:schema-structure}

We first show the schemas learned in one run of \emph{Blocks-3} and then compare the size and composition of the learned schemas with those of the original ones over all domains.

\paragraph{Example.}
\label{app:blocks-3-example}
\Cref{lst:blocks-3-diff} compares the original \emph{Blocks-3} operators with the operators that \ac{dias} learns in one run. The learned operators contain every original precondition and effect and add only negative preconditions, shown in green. Each of these is implied by the original preconditions in every reachable state, e.g., $\neg\mathtt{on}(x_0, x_4)$ by $\mathtt{on\text{-}table}(x_0)$ in \texttt{stack}, so the learned operators are equivalent to the original ones.

\paragraph{Size and composition.}
\Cref{tab:schema-sizes} compares the arity and the numbers of effects and preconditions of the schemas learned with action names with those of the original schemas, and \cref{tab:precondition-categories} splits the preconditions by kind. Both count distinct lifted literals, without the pairwise distinctness of parameters, which every learned schema states, and without the parameter types of the original schemas, which \ac{dias} learns as unary static preconditions. Of the 470 learned schemas (47 actions in 10 runs), 469 have as many effects as the original ones, all but one of \emph{Hanoi}, and 301 the same arity. In both \emph{Satellite} domains, the learned \texttt{calibrate} keeps only one of its three arguments. Positive fluent preconditions are close to the original ones, while most additional literals are negative fluent literals and literals over static relations.

\begin{table}[ht]
  \small
  \centering
  \caption{Original (GT) and learned (L) action schemas with action names, averaged over the actions of each domain and over 10 runs: arity, number of effects, and number of preconditions.}
  \label{tab:schema-sizes}
  \begin{tabular}{l c c cc c cc c cc}
    \toprule
    & & & \multicolumn{2}{c}{Arity} & & \multicolumn{2}{c}{Effects} & & \multicolumn{2}{c}{Preconditions} \\
    \cmidrule{4-5} \cmidrule{7-8} \cmidrule{10-11}
    Domain & Actions & & GT & L & & GT & L & & GT & L \\
    \midrule
    Blocks-3 & 3 & & 2.3 & 4.5 & & 3.3 & 3.3 & & 2.7 & 20.9 \\
    Delivery & 3 & & 3 & 3 & & 2.7 & 2.7 & & 2.3 & 41.7 \\
    Driverlog & 6 & & 3.2 & 3.1 & & 2.3 & 2.3 & & 2.3 & 66.8 \\
    Gripper & 3 & & 2.7 & 2.8 & & 2.7 & 2.7 & & 4.7 & 30.5 \\
    Hanoi & 1 & & 3 & 3.9 & & 4 & 4.2 & & 5 & 30 \\
    Logistics & 4 & & 3.2 & 4.3 & & 2 & 2 & & 5.8 & 85.3 \\
    Miconic & 4 & & 2 & 2.0 & & 2 & 2 & & 4 & 16.7 \\
    N-Puzzle & 4 & & 3 & 3 & & 4 & 4 & & 6 & 36 \\
    Satellite & 5 & & 2.8 & 2.5 & & 1.8 & 1.8 & & 2.8 & 57.7 \\
    Satellite-recal & 5 & & 2.8 & 2.3 & & 2 & 2 & & 3 & 43.8 \\
    Sokoban & 2 & & 2.5 & 4.6 & & 3 & 3 & & 4.5 & 50 \\
    Sokoban-pull & 3 & & 2.7 & 5.0 & & 3.3 & 3.3 & & 5 & 55.9 \\
    Spanner & 3 & & 3.3 & 3.7 & & 2.3 & 2.3 & & 3 & 71.9 \\
    Visitall & 1 & & 2 & 2 & & 3 & 3 & & 2 & 9 \\
    \bottomrule
  \end{tabular}
\end{table}

\begin{table}[ht]
  \small
  \centering
  \caption{Precondition literals of the original (GT) and learned (L) schemas with action names, summed over the actions of each domain, with learned counts averaged over 10 runs: unary and binary literals over static relations, and positive and negative literals over fluents.}
  \label{tab:precondition-categories}
  \begin{tabular}{l c cc c cc c cc c cc}
    \toprule
    & & \multicolumn{2}{c}{Static unary} & & \multicolumn{2}{c}{Static binary} & & \multicolumn{2}{c}{Positive fluent} & & \multicolumn{2}{c}{Negative fluent} \\
    \cmidrule{3-4} \cmidrule{6-7} \cmidrule{9-10} \cmidrule{12-13}
    Domain & & GT & L & & GT & L & & GT & L & & GT & L \\
    \midrule
    Blocks-3 & & 0 & 0 & & 0 & 0 & & 8 & 8.1 & & 0 & 54.7 \\
    Delivery & & 0 & 36 & & 1 & 27 & & 6 & 6 & & 0 & 56 \\
    Driverlog & & 0 & 93 & & 2 & 116.4 & & 12 & 11 & & 0 & 180.2 \\
    Gripper & & 8 & 25.2 & & 0 & 0 & & 6 & 6 & & 0 & 60.4 \\
    Hanoi & & 0 & 0 & & 2 & 11 & & 3 & 2.9 & & 0 & 16.1 \\
    Logistics & & 13 & 113 & & 2 & 76.1 & & 6 & 6 & & 2 & 146.2 \\
    Miconic & & 8 & 16.4 & & 2 & 17 & & 6 & 6 & & 0 & 27.4 \\
    N-Puzzle & & 12 & 24 & & 4 & 72 & & 8 & 8 & & 0 & 40 \\
    Satellite & & 0 & 50.8 & & 6 & 121.5 & & 8 & 4 & & 0 & 112.1 \\
    Satellite-recal & & 0 & 45.1 & & 6 & 86.1 & & 8 & 6 & & 1 & 81.8 \\
    Sokoban & & 0 & 0 & & 4 & 84.6 & & 3 & 3.2 & & 2 & 12.3 \\
    Sokoban-pull & & 0 & 0 & & 7 & 144.6 & & 5 & 5 & & 3 & 18.2 \\
    Spanner & & 0 & 55.5 & & 1 & 42.3 & & 8 & 9 & & 0 & 108.9 \\
    Visitall & & 0 & 2 & & 1 & 4 & & 1 & 2 & & 0 & 1 \\
    \midrule
    Total & & 41 & 461 & & 38 & 802.6 & & 88 & 83.2 & & 8 & 915.3 \\
    \bottomrule
  \end{tabular}
\end{table}

% |\lstgap{n}| at the end of a listing line leaves room for n unnumbered lines below it.
\newcommand{\lstgap}[1]{\vadjust{\vskip #1\baselineskip}}
\lstdefinestyle{pddldiff}{
  basicstyle=\ttfamily\footnotesize\selectfont,
  columns=fullflexible,
  keepspaces=true,
  showstringspaces=false,
  aboveskip=4pt,
  belowskip=0pt,
  moredelim=[is][\color{green!50!black}]{@}{@},
  moredelim=[is][\color{red}]{\#}{\#},
  numbers=left,
  numberstyle=\ttfamily\tiny\color{gray!50},
  numbersep=2em,
  xleftmargin=2.25em,
}

\begin{figure}[p]
  \captionof{lstlisting}{Original \emph{Blocks-3} schemas (left) and the schemas learned by \ac{dias} in one run (right). Learned variables are named after their slots, \texttt{?x0} to \texttt{?x4}. Differences are highlighted in red/green.}
  \label{lst:blocks-3-diff}
  \begin{minipage}[t]{0.48\linewidth}
\begin{lstlisting}[style=pddldiff, escapechar=|]
(:action stack
  :parameters (#?bm# #?bt#)
  :precondition (and
    (clear #?bm#)
    (clear #?bt#)
    (on-table #?bm#)
    (not (= #?bm# #?bt#))|\lstgap{4}|
  )
  :effect (and
    (not (clear #?bt#))
    (not (on-table #?bm#))
    (on #?bm# #?bt#)
  )
)

(:action newtower
  :parameters (#?bm# #?bf#)
  :precondition (and
    (clear #?bm#)
    (on #?bm# #?bf#)
    (not (= #?bm# #?bf#))|\lstgap{5}|
  )
  :effect (and
    (not (on #?bm# #?bf#))
    (on-table #?bm#)
    (clear #?bf#)
  )
)

(:action move
  :parameters (#?bm# #?bf# #?bt#)
  :precondition (and
    (clear #?bm#)
    (clear #?bt#)
    (on #?bm# #?bf#)
    (not (= #?bm# #?bt#))
    (not (= #?bm# #?bf#))
    (not (= #?bf# #?bt#))|\lstgap{10}|
  )
  :effect (and
    (not (clear #?bt#))
    (not (on #?bm# #?bf#))
    (on #?bm# #?bt#)
    (clear #?bf#)
  )
)
\end{lstlisting}
  \end{minipage}
  \hfill
  \begin{minipage}[t]{0.48\linewidth}
\begin{lstlisting}[style=pddldiff]
(:action stack
  :parameters (@?x0 ?x4@)
  :precondition (and
    (clear @?x0@)
    (clear @?x4@)
    (on-table @?x0@)
    (not (= @?x0 ?x4@))
    @(not (on ?x0 ?x0))@
    @(not (on ?x0 ?x4))@
    @(not (on ?x4 ?x0))@
    @(not (on ?x4 ?x4))@
  )
  :effect (and
    (not (clear @?x4@))
    (not (on-table @?x0@))
    (on @?x0 ?x4@)
  )
)

(:action newtower
  :parameters (@?x0 ?x2@)
  :precondition (and
    (clear @?x0@)
    (on @?x0 ?x2@)
    (not (= @?x0 ?x2@))
    @(not (clear ?x2))@
    @(not (on ?x0 ?x0))@
    @(not (on ?x2 ?x0))@
    @(not (on ?x2 ?x2))@
    @(not (on-table ?x0))@
  )
  :effect (and
    (not (on @?x0 ?x2@))
    (on-table @?x0@)
    (clear @?x2@)
  )
)

(:action move
  :parameters (@?x1 ?x2 ?x3@)
  :precondition (and
    (clear @?x1@)
    (clear @?x2@)
    (on @?x1 ?x3@)
    (not (= @?x1 ?x2@))
    (not (= @?x1 ?x3@))
    (not (= @?x2 ?x3@))
    @(not (clear ?x3))@
    @(not (on ?x1 ?x1))@
    @(not (on ?x1 ?x2))@
    @(not (on ?x2 ?x1))@
    @(not (on ?x2 ?x2))@
    @(not (on ?x2 ?x3))@
    @(not (on ?x3 ?x1))@
    @(not (on ?x3 ?x2))@
    @(not (on ?x3 ?x3))@
    @(not (on-table ?x1))@
  )
  :effect (and
    (not (clear @?x2@))
    (not (on @?x1 ?x3@))
    (on @?x1 ?x2@)
    (clear @?x3@)
  )
)
\end{lstlisting}
  \end{minipage}
\end{figure}

\subsection{Hanoi state-space coverage}
\label{app:hanoi-state-space}

\begin{figure}[!ht]
  \centering
  \caption{State graph of \emph{Hanoi} with six discs (\num{729} states). A random walk of \num{1500} steps from the initial state visits only \num{174} states along the black edges and stays within few sub-triangles of the Sierpiński structure.}
  \includegraphics[width=0.48\linewidth]{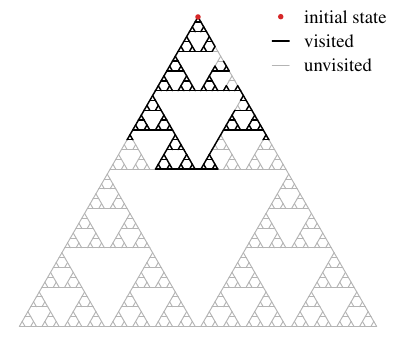}
  \label{fig:hanoi-walk}
\end{figure}

With $n$ discs and three pegs, the state graph $H_n$ of \emph{Hanoi} has $3^n$ states. It consists of three copies of $H_{n-1}$, one for each peg holding the largest disc, and each pair of copies is connected by a single edge that moves the largest disc. Recursively, $H_n$ forms a Sierpiński triangle, and a random walk can leave a sub-triangle only through two of its corner states, so it stays inside one for long stretches. \Cref{fig:hanoi-walk} shows this for six discs: a random walk of \num{1500} steps visits only \num{174} of the $3^6 = 729$ states. Our training instance has seven discs and $3^7 = \num{2187}$ states, of which the training walks of \num{1500} steps visit between 107 and 223 for seeds 0 to 9.

\subsection{Comparison with L1}
\label{app:l1-results}

We compare against the L1 algorithm of \citet{balyo:2024}. Its authors ran L1 on one random walk per domain on the training instance of \ac{dias} and returned the learned domains, which we evaluate once per domain with the same validator and test-state sampling as \ac{dias}. The resulting precision and recall values are summarized in \cref{tab:l1-results}.

\begin{table}[ht]
  \small
  \centering
  \caption{Results for L1~\citep{balyo:2024}. Precision and recall per domain, aggregated over the three test instances.}
  \label{tab:l1-results}
  \begin{tabular}{lcc}
    \toprule
    Domain & Precision & Recall \\
    \midrule
    Blocks-3 & 0.85 & 1.00 \\
    Delivery & 1.00 & 1.00 \\
    Driverlog & 0.35 & 1.00 \\
    Gripper & 1.00 & 1.00 \\
    Hanoi & 1.00 & 1.00 \\
    Logistics & 0.34 & 1.00 \\
    Miconic & 1.00 & 1.00 \\
    N-Puzzle & 1.00 & 1.00 \\
    Satellite & 0.66 & 1.00 \\
    Sokoban & 0.85 & 1.00 \\
    Sokoban-pull & 0.83 & 1.00 \\
    Spanner & 0.09 & 1.00 \\
    Visitall & 1.00 & 1.00 \\
    \bottomrule
  \end{tabular}
  
\end{table}

\subsection{Learning curves}
\label{app:learning-curves}

\Cref{tab:learning-curves} varies the amount of training data in five domains. Unlike in the other experiments, each run trains on exactly $N$ transitions per action name and validates on $N/2$, i.e., $N_\mathrm{min}^\mathrm{tr} = N_\mathrm{max}^\mathrm{tr} = N$ and $N_\mathrm{min}^\mathrm{val} = N_\mathrm{max}^\mathrm{val} = N/2$ in \cref{app:training-data}. Without noise, \emph{Blocks-3} and \emph{Gripper} are learned in every run from 20 transitions per action, and \emph{Delivery} from 50. In \emph{Hanoi}, no run is exact at 20 transitions, two from 50 on, and all three at \num{2000}. In \emph{Logistics}, every run but one is exact from 100 transitions on, and at most one below. At two flips, more data makes \emph{Delivery} exact in every run from 500 transitions on, but it does not help in the other domains. In \emph{Blocks-3}, every run is sound from 100 transitions on, but none is complete. In \emph{Logistics}, no run is sound at any size. \emph{Gripper} is exact in at most two runs at any size and \emph{Hanoi} in at most one.

\begin{table}[ht]
  \small
  \centering
  \caption{Learning curves with action names over seeds 0 to 2: number of sound and complete runs out of 3 when training on exactly $N$ transitions per action, without noise and at two flips per transition in expectation.}
  \label{tab:learning-curves}
  \addtolength{\tabcolsep}{-0.15em}
  \begin{tabularx}{\textwidth}{X c ccccccc c ccccccc}
    \toprule
    & & \multicolumn{7}{c}{No noise} & & \multicolumn{7}{c}{2 flips} \\
    \cmidrule{3-9} \cmidrule{11-17}
    $N$ & & 20 & 50 & 100 & 200 & 500 & 1000 & 2000 & & 20 & 50 & 100 & 200 & 500 & 1000 & 2000 \\
    \midrule
    Blocks-3 &  & 3 & 3 & 3 & 3 & 3 & 3 & 3 &  & 0 & 0 & 0 & 0 & 0 & 0 & 0 \\
    Delivery &  & 2 & 3 & 3 & 3 & 3 & 3 & 3 &  & 0 & 2 & 2 & 2 & 3 & 3 & 3 \\
    Gripper &  & 3 & 3 & 3 & 3 & 3 & 3 & 3 &  & 0 & 1 & 2 & 1 & 0 & 0 & 2 \\
    Hanoi &  & 0 & 2 & 2 & 2 & 2 & 2 & 3 &  & 1 & 0 & 0 & 0 & 0 & 0 & 1 \\
    Logistics &  & 1 & 0 & 3 & 3 & 2 & 3 & 3 &  & 0 & 0 & 0 & 0 & 0 & 0 & 0 \\
    \midrule
    Total (of 15) &  & 9 & 11 & 14 & 14 & 13 & 14 & 15 &  & 1 & 3 & 4 & 3 & 3 & 3 & 6 \\
    \bottomrule
  \end{tabularx}
\end{table}

\subsection{Component ablations}
\label{app:ablation-studies}
\label{app:component-ablations}

\Cref{tab:ablations} removes one component at a time with action names, over seeds 0 to 4 of \cref{tab:main-results}. Without slot weights, all slots are fully active, i.e., $\tilde{S}^a = S^a$. Without $L_\mathrm{pre}$ or $L_\mathrm{eff}$, the respective term is dropped from $L_\mathrm{aux}$, and without both, $\alpha = 0$. Without PCGrad, the two gradients are simply added, $g = g_\mathrm{main} + g_\mathrm{aux}$. Without random node features, all initial node features are zero.

Without $L_\mathrm{pre}$, 64 of the 70 runs are complete but only 12 sound, which is the permissive-precondition failure that this term prevents. Without PCGrad, the auxiliary terms override the prediction objective, and only 28 runs are sound and 5 complete. Without slot weights, every slot binds an object, and the schemas then require preconditions of objects that the action does not involve, which costs completeness in most domains (32 complete runs against 62). Removing $L_\mathrm{eff}$ mainly affects \emph{Logistics} and \emph{Spanner}, and the random node features matter least.

We do not ablate the \ac{gnn} encoder and the Sinkhorn assignment, since we see no informative variant. Without message passing, the keys equal the initial node features, so the selection no longer depends on the transition. A row-wise softmax in place of Sinkhorn lets one object occupy several slots, so the projection sums several cells of $P^a_\mathrm{eff}$ and $P^a_\mathrm{pre}$ into one entry of the adjacency matrix, which then no longer holds a probability. The argument-selection stage as a whole is isolated in \cref{tab:main-results}, where full STRIPS labels make it unnecessary and STRIPS+ labels fix part of it.

\begin{table}[ht]
  \small
  \centering
  \caption{Component ablations with action names over seeds 0 to 4. We report the number $N_{sc}$ of sound and complete runs out of 5 per domain, and the totals of $N_{sc}$, $N_s$, and $N_c$ over all 70 runs.}
  \label{tab:ablations}
  \begin{tabular}{l ccccccc}
    \toprule
    Domain & \ac{dias} & no $w^a$ & no $L_\mathrm{pre}$ & no $L_\mathrm{eff}$ & no $L_\mathrm{aux}$ & no PCGrad & no rand.\ init. \\
    \midrule
    Blocks-3 & 5 & 5 & 1 & 5 & 0 & 0 & 5 \\
    Delivery & 5 & 5 & 2 & 5 & 2 & 0 & 5 \\
    Driverlog & 5 & 0 & 0 & 5 & 0 & 0 & 4 \\
    Gripper & 5 & 0 & 3 & 5 & 2 & 0 & 5 \\
    Hanoi & 4 & 2 & 2 & 3 & 1 & 0 & 4 \\
    Logistics & 5 & 2 & 0 & 0 & 0 & 0 & 5 \\
    Miconic & 5 & 0 & 1 & 5 & 0 & 0 & 5 \\
    N-Puzzle & 5 & 4 & 1 & 5 & 0 & 0 & 5 \\
    Satellite & 0 & 0 & 0 & 0 & 0 & 0 & 0 \\
    Satellite-recal & 0 & 0 & 0 & 0 & 0 & 0 & 0 \\
    Sokoban & 2 & 0 & 0 & 1 & 0 & 0 & 1 \\
    Sokoban-pull & 5 & 5 & 0 & 5 & 0 & 5 & 4 \\
    Spanner & 5 & 2 & 0 & 0 & 0 & 0 & 5 \\
    Visitall & 5 & 5 & 2 & 4 & 2 & 0 & 5 \\
    \midrule
    $N_{s}$ (of 70) & 59 & 56 & 12 & 49 & 7 & 28 & 57 \\
    $N_{c}$ (of 70) & 62 & 32 & 64 & 58 & 63 & 5 & 59 \\
    $N_{sc}$ (of 70) & 56 & 30 & 12 & 43 & 7 & 5 & 53 \\
    \bottomrule
  \end{tabular}
\end{table}

\subsection{Sensitivity}
\label{app:sensitivity}

\Cref{tab:sensitivity} varies one hyperparameter at a time with action names, over seeds 0 to 4. In \emph{Blocks-3}, \emph{Gripper}, \emph{N-Puzzle}, and \emph{Sokoban-pull}, every $\alpha$ gives the same result, and the two \emph{Satellite} domains are not learned exactly at any $\alpha$. The other domains need different values: small ones in \emph{Driverlog}, large ones in \emph{Hanoi} and \emph{Spanner}, and intermediate ones in \emph{Logistics}, where small values drop the city parameter of \texttt{drive} and large ones over-constrain the schemas. We chose $\alpha$ for each domain in preliminary experiments, based on the validation results of a few runs, and not from this sweep. Variants of the same domain share one value. Except in \emph{Hanoi}, the chosen values (\cref{tab:domain-config}) yield as many sound and complete runs as any other value in the sweep. In \emph{Hanoi}, the chosen value $\alpha = 0.3$ gives 4 exact runs, against 5 with $\alpha = 1$. As training is deterministic, the runs at the chosen values coincide with seeds 0 to 4 of \cref{tab:main-results}. The best single value, $\alpha = 0.3$, recovers 49 of 70 runs, against 56 with the chosen per-domain values.

The derivation threshold matters little between 0.25 and 0.75, since most learned probabilities are close to 0 or 1. Without the schedule of $\tau$ ($T = 0$), $p_\mathrm{pre}$ is the plain product from the start, and 40 runs are exact. Values of $T$ from 100 to 1000 give 53 to 56 runs, while $T = 5000$ keeps $p_\mathrm{pre}$ close to the geometric mean for too long, and only 11 runs are exact. Seven slots instead of five cost four runs, and nine slots cost nine, five of them in \emph{Spanner}.

\begin{table}[ht]
  \small
  \centering
  \caption{Sensitivity with action names over seeds 0 to 4: number of sound and complete runs out of 5 per domain when varying one hyperparameter. $\alpha$: auxiliary-loss weight, with the value used for each domain in bold. Threshold: derivation threshold of \cref{sec:schema-derivation}, applied to the same trained models. $T$: steps until $\tau = 0.1$, i.e., $\tau = 0.1^{t/T}$. $M$: number of slots. The defaults are a threshold of 0.5, $T = 500$, and $M = 5$.}
  \label{tab:sensitivity}
  \addtolength{\tabcolsep}{-0.38em}
  \begin{tabularx}{\textwidth}{X c ccccc c ccccc c ccccc c ccc}
    \toprule
    & & \multicolumn{5}{c}{$\alpha$} & & \multicolumn{5}{c}{Threshold} & & \multicolumn{5}{c}{$T$} & & \multicolumn{3}{c}{$M$} \\
    \cmidrule{3-7} \cmidrule{9-13} \cmidrule{15-19} \cmidrule{21-23}
    Domain & & 0.01 & 0.03 & 0.1 & 0.3 & 1 & & 0.1 & 0.25 & 0.5 & 0.75 & 0.9 & & 0 & 100 & 500 & 1000 & 5000 & & 5 & 7 & 9 \\
    \midrule
    Blocks-3 &  & 5 & 5 & 5 & 5 & \textbf{5} &  & 5 & 5 & 5 & 5 & 5 &  & 5 & 5 & 5 & 5 & 5 &  & 5 & 5 & 5 \\
    Delivery &  & 5 & 5 & \textbf{5} & 5 & 4 &  & 4 & 5 & 5 & 5 & 5 &  & 5 & 5 & 5 & 5 & 0 &  & 5 & 5 & 5 \\
    Driverlog &  & \textbf{5} & 4 & 1 & 0 & 0 &  & 5 & 5 & 5 & 5 & 5 &  & 0 & 5 & 5 & 5 & 0 &  & 5 & 5 & 4 \\
    Gripper &  & 5 & 5 & 5 & 5 & \textbf{5} &  & 5 & 5 & 5 & 5 & 5 &  & 5 & 5 & 5 & 5 & 0 &  & 5 & 5 & 5 \\
    Hanoi &  & 3 & 2 & 2 & \textbf{4} & 5 &  & 4 & 4 & 4 & 4 & 4 &  & 4 & 5 & 4 & 4 & 4 &  & 4 & 4 & 3 \\
    Logistics &  & 0 & 0 & 5 & \textbf{5} & 0 &  & 2 & 2 & 5 & 5 & 3 &  & 0 & 5 & 5 & 5 & 0 &  & 5 & 5 & 4 \\
    Miconic &  & 5 & 4 & 5 & 5 & \textbf{5} &  & 5 & 5 & 5 & 5 & 5 &  & 5 & 5 & 5 & 5 & 1 &  & 5 & 5 & 5 \\
    N-Puzzle &  & 5 & 5 & 5 & 5 & \textbf{5} &  & 5 & 5 & 5 & 5 & 5 &  & 5 & 5 & 5 & 5 & 0 &  & 5 & 5 & 5 \\
    Satellite &  & \textbf{0} & 0 & 0 & 0 & 0 &  & 0 & 0 & 0 & 0 & 0 &  & 0 & 0 & 0 & 0 & 0 &  & 0 & 0 & 0 \\
    Satellite-recal &  & \textbf{0} & 0 & 0 & 0 & 0 &  & 0 & 0 & 0 & 0 & 0 &  & 0 & 0 & 0 & 0 & 0 &  & 0 & 0 & 0 \\
    Sokoban &  & 0 & 0 & \textbf{2} & 0 & 0 &  & 2 & 2 & 2 & 2 & 2 &  & 1 & 0 & 2 & 0 & 0 &  & 2 & 0 & 1 \\
    Sokoban-pull &  & 5 & 5 & \textbf{5} & 5 & 5 &  & 5 & 5 & 5 & 5 & 5 &  & 5 & 5 & 5 & 5 & 1 &  & 5 & 5 & 5 \\
    Spanner &  & 0 & 0 & 4 & 5 & \textbf{5} &  & 5 & 5 & 5 & 5 & 1 &  & 0 & 4 & 5 & 4 & 0 &  & 5 & 4 & 0 \\
    Visitall &  & 4 & 4 & 4 & 5 & \textbf{5} &  & 5 & 5 & 5 & 5 & 5 &  & 5 & 5 & 5 & 5 & 0 &  & 5 & 4 & 5 \\
    \midrule
    Total (of 70) &  & 42 & 39 & 48 & 49 & 44 &  & 52 & 53 & 56 & 56 & 50 &  & 40 & 54 & 56 & 53 & 11 &  & 56 & 52 & 47 \\
    \bottomrule
  \end{tabularx}
\end{table}

\subsection{Scalability}
\label{app:scalability}

\Cref{tab:scaling} reports training throughput and peak GPU memory for \emph{Blocks-3}, \emph{Logistics}, and \emph{Delivery} on instances with 12 to 150 objects. From 12 to about 150 objects, memory grows by a factor of 9 in \emph{Blocks-3} and 12 in \emph{Logistics}, and \emph{Delivery} runs out of memory at 150 objects. A step takes 2.3 times longer in \emph{Blocks-3}, 1.5 times in \emph{Logistics}, and 3.2 times in \emph{Delivery} at 105 objects. Memory is therefore the limiting resource. Since the learned schemas are lifted, however, they can be learned on small instances and apply unchanged to larger ones.

\begin{table}[ht]
  \small
  \centering
  \caption{Training cost by instance size on one NVIDIA A10 (\qty{24}{\giga B}): training steps per second (mean $\pm$ standard deviation) and peak allocated GPU memory, measured over 100 steps after 20 warm-up steps in five runs. OOM: out of memory.}
  \label{tab:scaling}
  \begin{tabular}{ll cccccc}
    \toprule
    Blocks-3 & Objects & 12 & 20 & 40 & 68 & 100 & 150 \\
    & Steps/s & 8.10{\scriptsize$\pm$0.52} & 8.06{\scriptsize$\pm$0.30} & 6.85{\scriptsize$\pm$0.13} & 5.59{\scriptsize$\pm$0.14} & 4.54{\scriptsize$\pm$0.16} & 3.53{\scriptsize$\pm$0.06} \\
    & GPU (GiB) & 0.7 & 1.0 & 1.8 & 2.9 & 4.1 & 6.2 \\
    \midrule
    Logistics & Objects & 12 & 21 & 38 & 67 & 103 & 147 \\
    & Steps/s & 4.22{\scriptsize$\pm$0.56} & 4.76{\scriptsize$\pm$0.57} & 4.51{\scriptsize$\pm$0.65} & 4.09{\scriptsize$\pm$0.29} & 3.45{\scriptsize$\pm$0.13} & 2.73{\scriptsize$\pm$0.07} \\
    & GPU (GiB) & 1.0 & 1.6 & 2.8 & 5.0 & 7.9 & 11.8 \\
    \midrule
    Delivery & Objects & 12 & 20 & 40 & 68 & 105 & 150 \\
    & Steps/s & 5.23{\scriptsize$\pm$0.80} & 4.15{\scriptsize$\pm$0.27} & 3.38{\scriptsize$\pm$0.21} & 2.27{\scriptsize$\pm$0.06} & 1.61{\scriptsize$\pm$0.03} & OOM \\
    & GPU (GiB) & 1.5 & 2.7 & 5.7 & 10.1 & 16.2 & OOM \\
    \bottomrule
  \end{tabular}
\end{table}

\IfFileExists{sections/checklist.tex}{%
  \clearpage
  \input{sections/checklist.tex}%
}{}

\end{document}